\documentclass[pdflatex,sn-mathphys-num]{sn-jnl}

\usepackage{comment}
\usepackage{subcaption}
\usepackage{tcolorbox}
\usepackage{pifont} 
\usepackage{amsmath}

\usepackage{graphicx}%
\usepackage{multirow}%
\usepackage{amsmath,amssymb,amsfonts}%
\usepackage{amsthm}%
\usepackage{mathrsfs}%
\usepackage[title]{appendix}%
\usepackage{xcolor}%
\usepackage{textcomp}%
\usepackage{manyfoot}%
\usepackage{booktabs}%
\usepackage{algorithm}%
\usepackage{algorithmicx}%
\usepackage{algpseudocode}%
\usepackage{listings}%

\theoremstyle{thmstyleone}%
\theoremstyle{thmstyletwo}%

\theoremstyle{thmstylethree}%

\begin{document}

\title[Article Title]{From Model to Classroom: Evaluating Generated MCQs for Portuguese with Narrative and Difficulty Concerns\footnote{\textbf{This is a preprint version of the manuscript currently under review at an international journal.}}}


\author[1]{\fnm{Bernardo} \sur{Leite}}\email{bernardo.leite@fe.up.pt}
\author[1]{\fnm{Henrique} \sur{Lopes Cardoso}}\email{hlc@fe.up.pt}

\author[2]{\fnm{Pedro} \sur{Pinto}}\email{papinto@portoeditora.pt}
\author[2]{\fnm{Abel} \sur{Ferreira}}\email{aferreira@portoeditora.pt}
\author[2]{\fnm{Lu\'{i}s} \sur{Abreu}}\email{labreu@portoeditora.pt}
\author[3]{\fnm{Isabel} \sur{Rangel}}\email{irangel@portoeditora.pt}

\author[4]{\fnm{Sandra} \sur{Monteiro}}\email{sandramonteiro@agrupamentoslourenco.org}

\affil[1]{
\orgdiv{LIACC}, \orgname{Faculdade de Engenharia, Universidade do Porto}, \orgaddress{\street{Rua Dr. Roberto Frias s/n}, \city{Porto}, \postcode{4200-465}, \country{Portugal}}}

\affil[2]{
\orgdiv{Departamento de Investigação e Tecnologia}, \orgname{Porto Editora},
\orgaddress{\street{Rua da Restaura\c{c}\~{a}o, 365}, \city{Porto}, \postcode{4099-023}, \country{Portugal}}}

\affil[3]{
\orgdiv{Divisão de Tecnologias Educativas}, \orgname{Porto Editora},
\orgaddress{\street{Rua da Restaura\c{c}\~{a}o, 365}, \city{Porto}, \postcode{4099-023}, \country{Portugal}}}

\affil[4]{
\orgdiv{Escola B\'{a}sica Mirante dos Sonhos}, \orgname{Agrupamento de Escolas de S\~ao Louren\c{c}o}, \orgaddress{\street{Rua Escola da Costa s/n}, \city{Ermesinde}, \postcode{4445-316}, \country{Portugal}}}


\abstract{Multiple-Choice Questions (MCQs) are among the most common assessment tools in education. 
While MCQs are valuable for learning and evaluation, manually creating them with varying difficulty levels and targeted reading skills remains a time-consuming and costly task.
Recent advances in generative AI provide an opportunity to automate MCQ generation efficiently. However, assessing the actual quality and reliability of generated MCQs has received limited attention -- particularly regarding cases where generation fails. This aspect becomes particularly important when the generated MCQs are meant to be applied in real-world settings. Additionally, most MCQ generation studies focus on English, leaving other languages underexplored.
This paper investigates the capabilities of current generative models in producing MCQs for reading comprehension in Portuguese, a morphologically rich language. Our study focuses on generating MCQs that align with curriculum-relevant narrative elements and span different difficulty levels. We evaluate these MCQs through expert review and by analyzing the psychometric properties extracted from student responses to assess their suitability for elementary school students.
Our results show that current models can generate MCQs of comparable quality to human-authored ones.
However, we identify issues related to semantic clarity and answerability.
Also, challenges remain in generating distractors that engage students and meet established criteria for high-quality MCQ option design.
}




\maketitle

\section{Introduction}\label{sec1}

Objective questions~\citep{das_2021_objective}, such as multiple-choice (MCQs), require test-takers to select the correct answer from a list of options, where the remaining choices serve as distractors or incorrect alternatives~\citep{kurdi_2020_education}. These questions are widely used in education as they promote standardized evaluation and allow educators to assess large groups of students efficiently while minimizing bias~\citep{ch_2020_mcq_survey}.
However, crafting high-quality MCQs manually is a demanding process. It involves carefully designing distractors that are misleading enough to challenge students but still contribute to meaningful learning~\citep{alhazmi_2024_distractor_survey}. 

To this end, the research community has consistently advanced automatic Question Generation (QG), particularly MCQ generation, across various domains.
This joint effort is driven by the assumption that automating question creation not only reduces the manual workload for educators but also enhances the scalability of assessment activities~\citep{das_2021_objective}. However, despite its potential, the actual adoption of these generation models in educational platforms remains unclear, alongside the technical challenges that must be addressed when integrating QG systems into real-world scenarios~\citep{lee_2024_few_shot_is_enough}. In fact, as reported by \citet{wang_2022_towards_modular}, automatic QG tools are not widely adopted in classrooms.

We believe that the low adoption of QG tools is linked to two key issues that have been previously analyzed in the literature and motivate our research.
First, since the classroom is a high-stakes environment and instructors often have predefined goals, teachers may be reluctant to rely on imperfect and fallible AI models \citep{holstein_aleven_2022_fallible_ai,wang_2022_towards_modular}.
Second, the reliability of generated MCQs has received limited attention, particularly regarding instances where generative models fail to produce accurate questions, making these systems untrustworthy. Other concerns suggest that generated questions perform well only in narrowly defined domains, and are limited in types and difficulty levels~\citep{kurdi_2020_education,alsubait_2016_ontology,wang_2022_towards_modular}.
For instance, while some studies assert that generated medical MCQs using ChatGPT\footnote{\url{https://openai.com/index/chatgpt/}} exhibited acceptable levels~\citep{kiyak_2024_chatgpt_medical}, others highlight critical errors associated with generating MCQs about vocabulary~\citep{malec_2024_csedu_mcq_vocabulary}.

Naturally, MCQ generation performance also depends on the target language. Despite continuous interest in this field, most research on MCQ generation focuses on English, leaving other languages underexplored---largely due to limited resources~\citep{alhazmi_2024_distractor_survey}. 
Portuguese is one such language, recognized for its morphological richness, with complex inflectional and derivational structures.

In this study, we investigate the capability of state-of-the-art generative models to generate reading comprehension MCQs for Portuguese elementary students.
Difficulty and narrative-controllable QG plays a crucial role in education, as it facilitates the creation of personalized questions that address students' unique needs and learning goals~\citep{kurdi_2020_education}.
Aligned with this objective, we examine the ability of current models to generate MCQs with different difficulty levels and curriculum-relevant narrative elements.

Our evaluation involves both quantifying and qualifying the performance of the most up-to-date models (despite the rapid advancements in this field) in generating appropriate MCQs for elementary students. This assessment is based on two main perspectives: expert review and psychometric properties. The expert review focuses on evaluating the reliability of the generated MCQs based on specific quality metrics.
For the psychometric properties, we evaluate the validity of MCQs by analyzing actual student responses.

Our results show that generated MCQs are of comparable quality to human-authored ones.
We identified issues such as semantic clarity in questions and problems related to answerability, but these are present in both human-authored and generated MCQs.
Human-authored MCQs are perceived as slightly more difficult and more discriminative. Additionally, they tend to engage students more effectively and adhere more closely to psychometric best practices for option selection.

While this study is naturally constrained by the performance of the applied models for Portuguese, any identified limitations can be generalized to other contexts and languages, particularly Romance languages.

The remainder of this paper is organized as follows. 
Section~\ref{sec:methods_mcq_gen} presents the two methods used for generating MCQs.
Section~\ref{sec:eval_expert} describes the expert review process.
Section~\ref{sec:eval_students} evaluates the MCQs using psychometric properties derived from student responses.
Section~\ref{sec:perceived_difficult} examines the perceived difficulty of the generated MCQs from multiple perspectives.
Section~\ref{sec:summary_findings} summarizes the main findings, and Section~\ref{sec:limitations} outlines the study's limitations.
Finally, Section~\ref{sec:conclusions} concludes the study.

\section{Background \& Related Work}

\subsection{Purpose and Value of Generated MCQs} \label{sec:purpose_of_mcqs}

In this study, we aim at generating \textit{reading comprehension} MCQs.
A typical example of a reading comprehension MCQ is illustrated in Figure~\ref{fig:mcq_example}. In this case, a narrative text serves as the input for generating a \textit{wh}-question\footnote{An open-ended question that begins with words like who, what, where, when, why, or how --- commonly referred to as ''wh-questions'' in English. In our case, this forms the question part of a complete MCQ, followed by one correct answer and distractors.}, a correct option, and three distractors.

\begin{figure}[!ht]
    \centering
    \begin{tcolorbox}[width=0.8\textwidth, colframe=black, colback=gray!10, arc=2mm]
        \textbf{Narrative text:}  
        
        Um dia, porém, o elefantezinho cor-de-rosa sentiu uma esquisita sensação, quando viu que uma flor branca murchava... A flor ia morrer! Aflito, chamou os companheiros que vieram...
        {\tiny (One day, however, the little pink elephant felt a strange sensation when he saw that a white flower was wilting... The flower was going to die! Distressed, he called out to his companions, who came...)}  
        
        \vspace{0.3cm}
        \textbf{Generated MCQ:}  
        
        Como se sentiu o elefantezinho cor-de-rosa ao ver a flor murchar?
        {\tiny (How did the little pink elephant feel when he saw the flower wither?)}  
        
        \vspace{0.3cm}
        
        A) Tranquilo. {\tiny (Relax.)}
        
        B) Aflito. \ding{51} {\tiny (Distressed.)}  
        
        C) Feliz. {\tiny (Happy.)}  
        
        D) Indiferente. {\tiny (Indifferent.)}  
    \end{tcolorbox}
    \caption{Illustrative example of a generated MCQ.}
    \label{fig:mcq_example}
\end{figure}

We focus on reading comprehension since it is a skill that impacts multiple aspects of education, including language acquisition and cognitive development~\citep{ashok_2023_rc}. Furthermore, the ability to comprehend and interpret stories, such as fairy tales, plays a key role in fostering early intellectual and literacy development in children~\citep{sim_2014_rc,lynch_2008_rc}.

Specifically, our approach aligns with the pedagogical goal established for the Portuguese (native) language curriculum at the elementary school level, emphasizing \textit{the comprehension of narrative and descriptive texts, including the analysis and questioning of key narrative elements}\footnote{This pedagogical goal has been taken from the established essential learning skills: \url{https://www.dge.mec.pt/aprendizagens-essenciais-ensino-basico}}.
As reported by \citet{xu_2022_fairytaleqa}, narrative comprehension is a high-level cognitive skill closely linked to overall reading proficiency~\citep{lynch_2008_rc}. Therefore, this study aims to generate MCQs that specifically target different types of narrative elements during the generation process, including \textit{character}, \textit{feelings}, \textit{setting}, \textit{action}, and \textit{causal relationships} (see Appendix~\ref{sec:appendix_narratives}). Additionally, controlling question difficulty allows for a more balanced assessment of children's reading comprehension skills~\citep{eo_2023_diversity,xu_2022_fairytaleqa}. To address this, our study also focuses on generating MCQs with varying levels of difficulty.

\subsection{Research on MCQ Generation}
From a methodological perspective, research on MCQ generation initially relied on feature-based approaches, which leverage corpus features~\citep{alhazmi_2024_survey_mcq}. For instance, the selection of distractors has been based on word frequency~\citep{jiang_2017_dist}, part-of-speech (POS) tagging~\citep{susanti_2015_dist}, morphological characteristics~\citep{pino_2009_dist}, or co-occurrence~\citep{jiang_2017_dist} with the correct answer. Additionally, selecting distractors based on their syntactic, semantic or contextual similarity to the correct answer has been a common technique~\citep{kurdi_2020_education}.

With the advancement of neural-based methods, research began adopting encoder-decoder models. Notably, the RACE dataset~\citep{lai_2017_race} has served as a popular benchmark for this purpose~\citep{gao_2019_race,zhou_2020_race}.
Building on this, models such as T5~\citep{raffel_2020_t5} and BART~\citep{lewis_2020_bart}, which are Transformer-based~\citep{vaswani_2017_transformer} models pre-trained on large-scale corpora, have been fine-tuned on smaller, domain-specific datasets to accomplish the MCQ generation task~\citep{rodriguez_2022_mcq,wang_2023_dist,taslimipoor_2024_dist}.

More recently, with the emergence of large language models (LLMs), prompt engineering has been used to instruct these LLMs to perform specific tasks, including MCQ generation with zero and few-shot strategies~\citep{bitew_2025_fewshot,lee_2024_few_shot_is_enough,lin_2024_psyco}. 
While prompt engineering with LLMs has demonstrated remarkable performance in general text generation tasks, there remains a lack of in-depth empirical studies and validation efforts assessing their effectiveness in generating MCQs with real-world applicability in educational settings~\citep{lee_2024_few_shot_is_enough,lin_2024_psyco}. To address this, our research explores zero-shot prompting techniques with LLMs for this type of analysis. For comparison purposes, we also investigate a fine-tuned model with a more modest parameter size ($<$ 1 billion).

Regarding MCQ evaluation, ranking-based metrics are commonly used to assess models by measuring the relevance of the top-\textit{k} retrieved distractors. Also, \textit{n}-gram metrics, such as BLEU~\citep{papineni_bleu_2002} and ROUGE~\citep{lin_rouge_2004}, evaluate the lexical overlap between generated and reference (human-authored) distractors. Human evaluation, which focuses on aspects such as reliability, plausibility, and difficulty, is also widely applied.
In this study, we first conduct an expert review evaluation process. In addition to commonly used human evaluation metrics, we assess the narrative alignment of the generated MCQs and their difficulty for the target population. Second, a key contribution of this research is the application of pre-validated MCQs in real student assessments.

To the best of our knowledge, MCQ generation in Portuguese has previously been explored only through rule-based techniques~\citep{curto_mthesis_2010,correia_mcqpt_2010,leite_2020_msc} and with small models for distractor ranking~\citep{oliveira_mcqpt_2023}.

\subsection{Controllable Question Generation}

Controlling the types of generated questions has been a focus point in automatic QG~\citep{kurdi_2020_education}, under the goal of creating tailored questions that accommodate students' diverse needs and learning objectives.
For instance, \citet{ghanem_2022_cqg_acl} proposed controlling the generation process to focus on specific reading comprehension skills, such as figurative language and vocabulary.
\citet{elkins_2023_cqg_aied} propose to control Bloom's question taxonomy \citep{bloom_2002_revised}. 
More relevant to this study, research on narrative-controlled QG has gained attention, largely due to the availability of FairytaleQA~\citep{xu_2022_fairytaleqa}, a high-quality dataset designed by education experts where each question is annotated with a narrative label. Some studies~\citep{zhao_2022_cqg_acl,leite_2023_cqg_aied} have approached this control mechanism by fine-tuning encoder-decoder models with control labels, while others have adopted prompt engineering through few-shot strategies with LLMs~\citep{leite_2024_fcqg_csedu,zhang_2024_plans}.
While previous studies have primarily focused on controlling the generation of open-ended \textit{wh}-questions, mainly due to the structure of FairytaleQA, this study extends narrative control to MCQ generation, influencing not only the \textit{wh}-question itself but also the answer options.

Difficulty control is more related to the challenge posed on answering the generated questions correctly. For open-ended \textit{wh}-questions, \citet{yifangao_2019_dqg} assigned difficulty labels (easy or hard) to questions based on whether QA systems could answer them correctly, then used these labels as inputs to control the generation process. \citet{kumar_2019_dcqg} proposed estimating difficulty based on named entity popularity, while \citet{bi_2021_dcqg} tackle the challenge of high diversity in QG. Furthermore, \citet{cheng_2021_dcqg} controlled question difficulty by considering the number of inference steps required to arrive at an answer.
Notably, recent studies have explored the relationship between question difficulty and the learner's ability~\citep{uto_2023_dif,tomikawa_2024_adpative}.

For MCQs, the primary approach to addressing difficulty has been to compute the degree of similarity between the correct answer and distractors~\citep{lin_2015_dif,alsubait_2016_ontology,kurdi_2016_dif,susanti_2017_cid}, whether this similarity is feature-based, syntactic, or semantic.
More recently, \citet{tomikawa_2024_difmcqg} proposed an approach where responses from multiple QA systems were used to annotate MCQs from the RACE dataset with difficulty values. A model was then trained on these annotations to learn how to control difficulty.
However, following such an approach requires the availability of large-scale datasets. For instance, RACE contains nearly 100k MCQs, but in real educational settings, particularly in lower-resourced languages, these datasets are not available. Additionally, it remains uncertain whether (simulated) QA systems can accurately mimic student response behavior.
Finally, difficulty is often a subjective concept, as it can vary depending on the respondent.

In this study, rather than proposing new features or methods for controlling difficulty, we aim to investigate how generative models perceive difficulty when instructed to assign difficulty values to their generated MCQs. During evaluation, we compare model-assigned difficulty values with expert review and psychometric properties (actual student performance).

\subsection{This Study: Research Questions}

Based on the above review, this study aims to investigate the capability of current generative models to generate reading comprehension MCQs for Portuguese elementary students, evaluating their quality through expert review and psychometric models. The research questions (RQs) are as follows:

RQ1: How acceptable are MCQs generated by current generative models?

RQ2: How do the psychometric properties of generated MCQs compare to those of human-authored ones?

RQ3: How do the perceived difficulty characteristics of generated MCQs compare across expert review, psychometric properties, and model-annotated difficulty values?

\section{Method for MCQ Generation} \label{sec:methods_mcq_gen}

This section presents two methods for generating MCQs.
In the first method (Section~\ref{sec:method_one_step}), MCQs are generated in a single step: we prompt the model to generate five MCQs simultaneously. Each MCQ includes a question, correct answer, distractors, difficulty level, and narrative label.
In the second method (Section~\ref{sec:method_two_step}), MCQs are generated in two steps: first, a model generates one \textit{wh}-question conditioned on a single narrative label; then, a second model generates the remaining components (distractors and difficulty level). This process is repeated five times to produce five MCQs, as in the first method.

\subsection{One-Step MCQ Generation} \label{sec:method_one_step}
\begin{figure}[!ht]
    \centering
    \includegraphics[width=\textwidth]{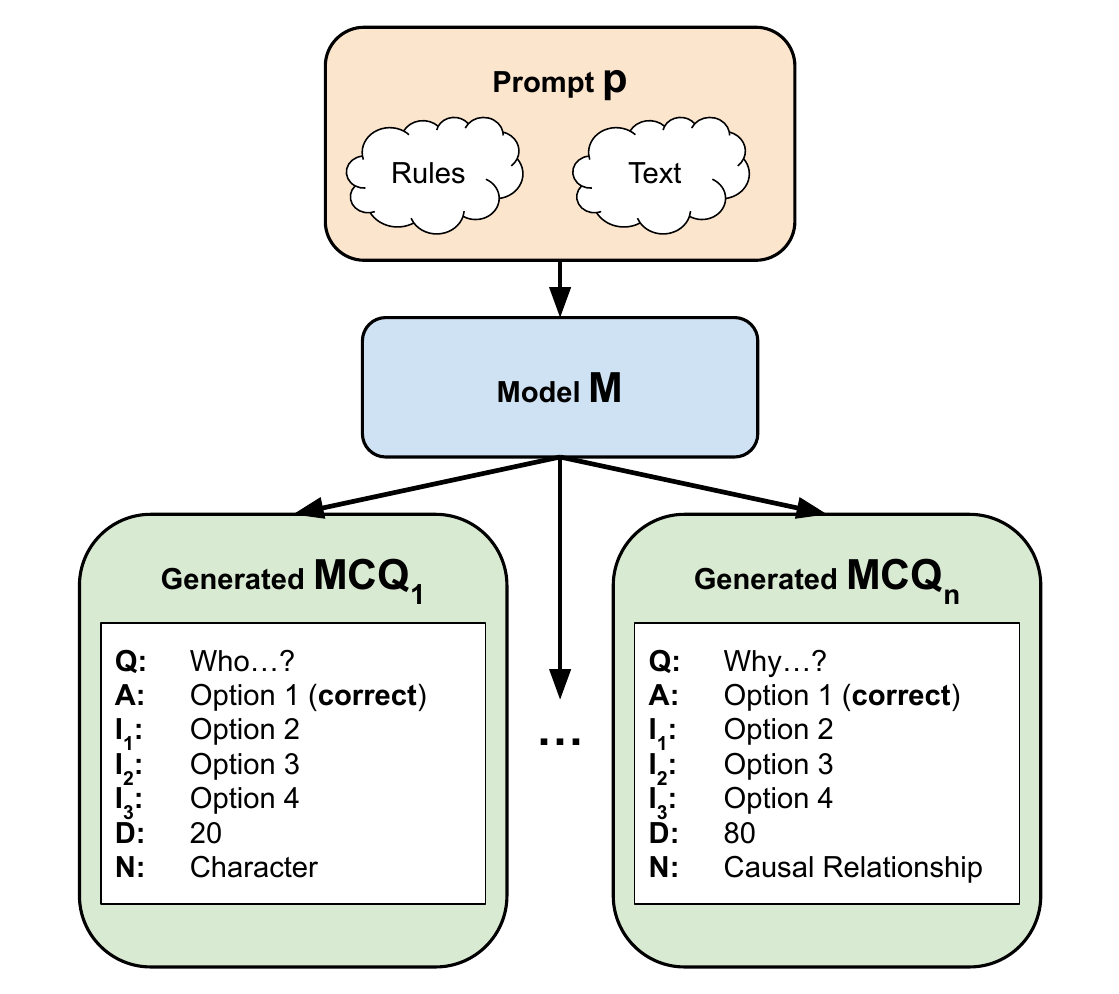}
    \caption{One-step method for generating MCQs.}
    \label{fig:one_step_mcqg}
\end{figure}

In one-step MCQ generation, zero-shot prompting is employed with LLMs.
Figure~\ref{fig:one_step_mcqg} shows the overall process.

Specifically, given an instruction prompt \( p \), the objective is to use a model \( M \) to generate a set of \( MCQs \). This process can be formally defined as:
\begin{equation}
MCQs = M(p),
\end{equation}
where prompt $p$ includes the target text, task description, generation rules, and the expected output format. The model \( M \) is a generative decoder-only model. The generated output consists of a set of $MCQs$, denoted as \( MCQ_1, \dots, MCQ_n \). Each generated $MCQ$ can be formally defined as:
\begin{equation}
MCQ: \langle Q, A, I_1, I_2, I_3, D, N \rangle
\end{equation}
where \( Q \) is the \textit{wh}-question, \( A \) is the correct answer, \( I_1, I_2, I_3 \) are three incorrect answers (distractors), \( D \) represents the difficulty value assigned by the model, and \( N \) is the narrative label assigned by the model. This approach ensures that each MCQ includes both a difficulty value and a narrative element, aligning with our goal of controlled MCQ generation.
The process of designing an effective prompt for MCQ generation requires applying prompt engineering principles alongside iterative testing and refinement~\citep{ray_principles_2023,lin_2024_psyco}. Therefore, the prompt development for this task followed an \textit{interactive prompting} approach~\citep{heston_interactiveprompt_2023}, based on \citet{lin_2024_psyco}, which can be divided into three key stages: (1) defining the initial prompt, (2) establishing generation rules and output format, and (3) conducting overall testing and finalizing the prompt.

\subsubsection{Initial prompt definition}
The first step involved drafting an initial prompt for MCQ generation and conducting preliminary analyses. At this stage,
the model was instructed to generate five MCQs from a given text. The number five was chosen based on the number of narrative elements analyzed (recall Section~\ref{sec:purpose_of_mcqs}), allowing the model to generate one MCQ per element.

\subsubsection{Establishing generation rules} \label{sec:generation_rules}
Specific constraints were introduced to guide the generation process. The MCQs were required to be written in European Portuguese, ensuring linguistic and cultural appropriateness. Additionally, we enforced rules such as a structured MCQ output format, the presence of a single correct answer, and distractors that are contextually relevant; that is, they are incorrect yet meaningfully related to the text. The MCQs were also tailored for children around the age of eight. This age was selected as a developmental reference where children possess fundamental literacy skills. Since eight years old falls in the middle of Portuguese elementary school (ages 6–10), this choice attempts to ensure accessibility to a broad range of students while maintaining an appropriate level of challenge.

\subsubsection{Overall testing and finalization} \label{sec:final_prompt}

The final stage involved evaluating the prompt's effectiveness through multiple trials, refining it based on observed outputs, and selecting the most stable version for MCQ generation. The final prompt was originally written in Portuguese, and its English translation is shown in Appendix~\ref{sec:full_prompt_image}.

\subsection{Two-Step MCQ Generation} \label{sec:method_two_step}

\begin{figure}[!ht]
    \centering
    \includegraphics[width=\textwidth]{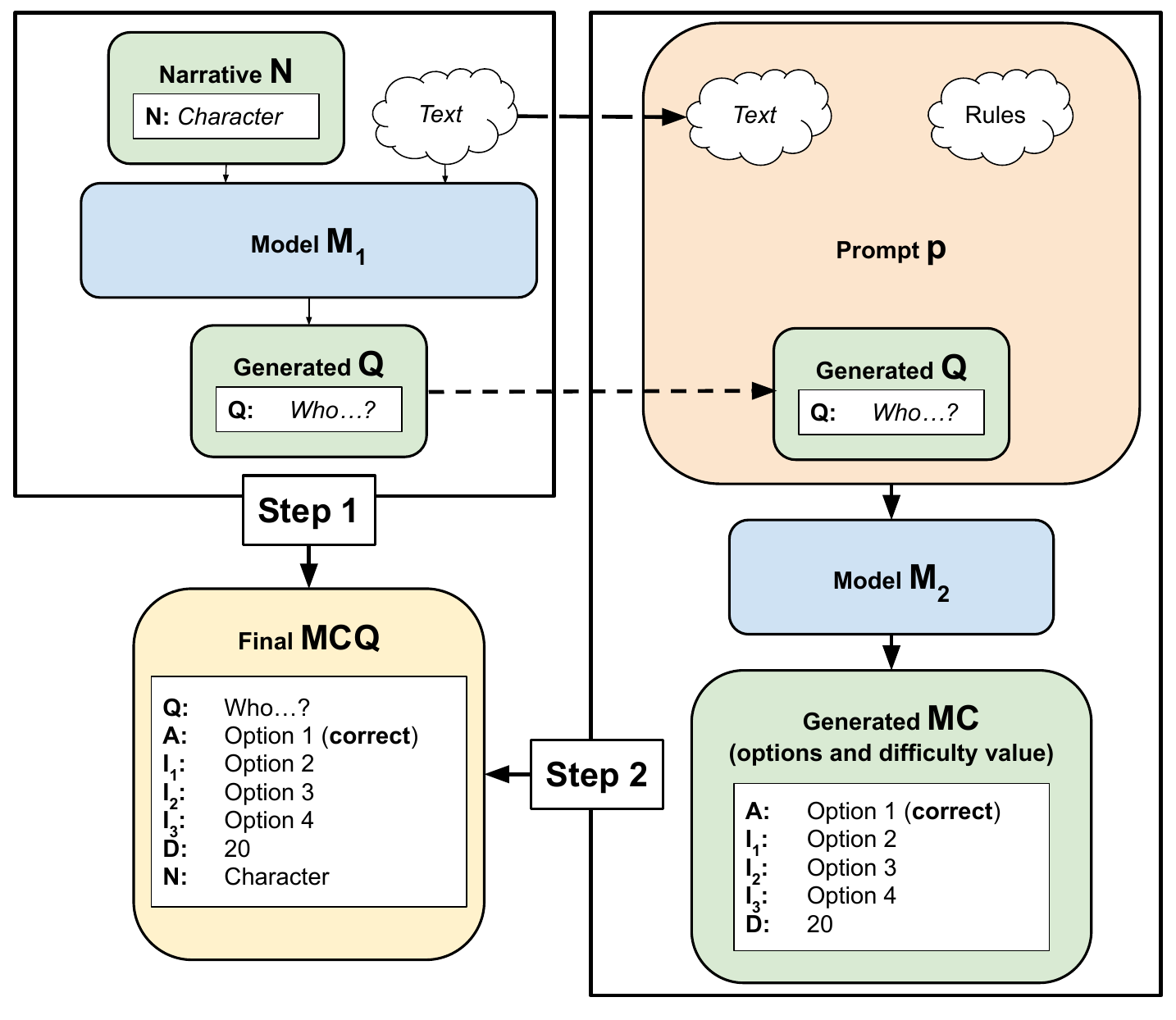}
    \caption{Two-step method for generating MCQs.}
    \label{fig:two_step_mcqg}
\end{figure}

In two-step MCQ generation, we first use one model to generate the \textit{wh}-question and then another model to generate the remaining components of the MCQ.
Figure~\ref{fig:two_step_mcqg} shows the overall process.
This method is motivated by the premise that modular QG with distinct components can offer greater flexibility than generating the entire output in a single step~\citep{wang_2022_towards_modular}. For instance, it allows the possibility of using only the \textit{wh}-question without necessarily proceeding to the generation of a full MCQ.
Additionally, dividing the NLP tasks into sequential steps and chaining the outputs has been reported as a way to improve the quality of task outcomes~\citep{wu_2022_divide_nlp_tasks,wang_2022_towards_modular}.
Formally, the \textbf{first step} can be represented as follows:
\begin{equation}
Q = M_{1}(T,N)
\end{equation}
where the text \( T \) and target narrative label \( N \) are provided as input, and the goal is to generate a \textit{wh}-question \( Q \) using \( M_1 \): 
an encoder-decoder model fine-tuned with appropriate data (described in Section~\ref{sec:method_details}). The encoder processes the text and narrative label, encoding them into a fixed-length representation known as a context vector. The decoder then takes this context vector and generates the output \textit{wh}-question \( Q \). The objective is to guide the model in producing a \textit{wh}-question \( Q \) that aligns with the intended narrative element \( N \).
The \textbf{second step} is then formalized as follows:
\begin{equation}
MC = M_{2}(p),
\end{equation}
where the prompt \( p \) already includes the \textit{wh}-question \( Q \) from Step 1, along with the target text and generation rules. The model \( M_{2} \) is a generative decoder-only model.
Each generated \( MC \) is formally defined as:
\begin{equation}
MC: \langle A, I_{1}, I_{2}, I_{3}, D \rangle,
\end{equation}
where \( A \) is the correct answer and \( I_1, I_2, I_3 \) are the three incorrect answers. The value \( D \) denotes the difficulty value assigned by model \( M_{2} \).
The final \( MCQ \) is obtained by merging the components from steps 1 and 2:
\begin{equation}
MCQ: \langle Q, A, I_{1}, I_{2}, I_{3}, D, N \rangle.
\end{equation}
This process (steps 1 and 2) is repeated five times using the same text while changing the narrative label to one of the five studied narrative elements. By doing this, five MCQs per text are obtained, similar to the One-Step MCQ method (Section~\ref{sec:method_one_step}). 
The task description and generation rules undergo only subtle modifications from the prompt used in the One-Step MCQ method to accommodate the process of generating the MCQ based on the pre-generated \textit{wh}-question.

\subsection{Model Choice and Experimental Details} \label{sec:method_details}
For One-Step MCQ Generation, we use \texttt{GPT-4o}\footnote{Latest version as of October 2024.}. Given its multilingual text understanding and generation capabilities, including in Portuguese, we consider it essential for our analysis.
Additionally, our goal was to explore an alternative open-weight, multilingual model with Portuguese generation capability for comparative purposes. To this end, we conducted a small comparative experiment with some models.
We selected \texttt{Gemma-2-27B} as the model that demonstrated the best performance.

For Two-Step MCQ Generation, we use the open-source \texttt{ptt5-v2-large}\footnote{\url{https://huggingface.co/unicamp-dl/ptt5-v2-large}}~\citep{piau_2024_ptt5v2} model to generate the \textit{wh}-question (first step). It is a modest-scale (740M) encoder-decoder model that we fine-tuned on the European Portuguese machine-translated version of FairytaleQA\footnote{\url{https://huggingface.co/datasets/benjleite/FairytaleQA-translated-ptPT}}~\citep{leite_2024_fairy_pt}. The dataset includes questions annotated with multiple narrative labels, including the five used in this study.
In the second step, which involves generating the remaining MCQ components, we use again \texttt{Gemma-2-27B}.
To improve readability in the following sections, from now on, we will refer to these models as follows:
\begin{itemize}
    \item \texttt{GPT-4o} (One-Step MCQ Generation)
    \item \texttt{Gemma-2} (One-Step MCQ Generation)
    \item \texttt{Ptt5-v2+Gemma-2} (Two-Step MCQ Generation)
\end{itemize}

\section{Evaluation: Expert Review} \label{sec:eval_expert}

This section describes the expert evaluation process, covering participant characterization, data selection, evaluation procedure, metrics and results obtained.

\subsection{Participants}

We engaged 18 native Portuguese-speaking participants with higher education degrees who have professional experience related to Portuguese language education for young children. Their expertise includes teaching, developing and managing educational learning resources. Of these, 10 are collaborators of \textit{Porto Editora}\footnote{\url{www.portoeditora.pt/}}, Portugal's largest publishing group, which supported this study. The remaining 8 are teachers who voluntarily participated in the evaluation.

\subsection{Data Collection}

We manually selected 12 narrative texts, all recommended by Portugal's National Reading Plan\footnote{\url{https://pnl2027.gov.pt/np4EN/file/33/NRPFramework.pdf}}. Elementary education in Portugal spans four years, with students in the first year still developing reading and writing skills. Therefore, we excluded first-year texts and selected 4 texts for each of the 2nd, 3rd, and 4th years, for a total of 12 texts.
All texts are sourced from the \textit{Escola Virtual}\footnote{\url{www.escolavirtual.pt/}} online platform, owned by \textit{Porto Editora}, which provides a wide range of educational resources for students of all ages, including elementary school students. Alongside the texts, we also collected the corresponding human-authored reading comprehension MCQs, which serve as a benchmark for comparison with the generated MCQs.

\subsection{Evaluation Procedure}

The evaluation protocol can be summarized as follows:

\begin{enumerate}
    \item \textbf{MCQs Generation}: Five MCQs are generated for each of the 12 narrative texts. Since we use three models (recall Section~\ref{sec:method_details}), we obtain a total of 180 generated MCQs.
    \item \textbf{Collecting Human-Authored MCQs}: Human-authored MCQs are added to the generated ones. We collected 58 human-authored MCQs, resulting in a total of 238 MCQs.
    \item \textbf{Building 12 Forms}: We create 12 Google Forms\footnote{\url{https://workspace.google.com/products/forms/}}, one for each narrative text, including the text and its corresponding MCQs. Each form includes 15 MCQs, randomly extracted from both the human-authored and the generated pools. To ensure balanced and manageable evaluation, a final subset of 180 MCQs is selected: 45 human-authored, 45 from \texttt{GPT-4o}, 45 from \texttt{Gemma-2}, and 45 from \texttt{Ptt5-v2+Gemma-2}.
    \item \textbf{Form Distribution}: The 12 forms are randomly assigned to the 18 participants. Depending on availability, each participant is given a minimum of one and a maximum of three forms. We ensure that each form --- and consequently, each MCQ --- is evaluated by exactly three participants, enabling majority voting.
    \item \textbf{Expert Review}: For each form, participants are instructed to read the assigned narrative text and rate the set of 15 MCQs (without knowing their provenance) based on appropriate metrics (described in Section~\ref{sec:human_eval_metrics}).
\end{enumerate}

The distribution of forms is shown in Table~\ref{tab:questions_per_form}. This evaluation took place between November 11 and December 6, 2024.

\begin{table}[h]
    \caption{Form contents and assigned participants: each form contains a text, 15 MCQs from four provenances (\texttt{Human}, \texttt{GPT-4o}, \texttt{Gemma-2}, and \texttt{Ptt5-v2+Gemma-2}), and is assigned to three participants.}
    \label{tab:questions_per_form}
    \centering
    \begin{tabular*}{\textwidth}{@{\extracolsep\fill}lcc ccccc c}
        \toprule
        Form ID & Participants & Text Year & \texttt{Human} & \texttt{GPT-4o} & \texttt{Gemma-2} & \texttt{Ptt5-v2+Gemma-2} & Total \\
        \midrule
        1  & $\langle$A, B, C$\rangle$  & 2nd  & 3 & 3 & 4 & 5 & 15 \\
        2  & $\langle$A, B, C$\rangle$  & 2nd  & 3 & 3 & 4 & 5 & 15 \\
        3  & $\langle$A, B, C$\rangle$  & 2nd  & 4 & 3 & 3 & 5 & 15 \\
        4  & $\langle$D, E, F$\rangle$  & 2nd  & 3 & 5 & 4 & 3 & 15 \\
        5  & $\langle$D, E, F$\rangle$  & 3rd  & 3 & 4 & 5 & 3 & 15 \\
        6  & $\langle$D, E, F$\rangle$  & 3rd  & 4 & 3 & 5 & 3 & 15 \\
        
        7  & $\langle$G, H, I$\rangle$  & 3rd  & 4 & 3 & 3 & 5 & 15 \\
        8  & $\langle$G, H, J$\rangle$  & 3rd  & 5 & 4 & 3 & 3 & 15 \\
        9  & $\langle$G, H, K$\rangle$  & 4th  & 5 & 4 & 3 & 3 & 15 \\
        
        10 & $\langle$L, M, N$\rangle$  & 4th  & 4 & 5 & 3 & 3 & 15 \\
        11 & $\langle$L, O, P$\rangle$  & 4th  & 3 & 5 & 3 & 4 & 15 \\
        12 & $\langle$L, Q, R$\rangle$  & 4th  & 4 & 3 & 5 & 3 & 15 \\
        \midrule
        Total   & 18  &  & 45 & 45 & 45 & 45 & 180 \\
        \botrule
    \end{tabular*}
\end{table}

\subsection{Evaluation Metrics} \label{sec:human_eval_metrics}

First, participants were instructed to assess each item based on (1) \textbf{well-formedness}, determining whether the \textit{wh}-question is well-formed, free of orthographic, grammatical, and semantic errors. 
We then proceed to (2) \textbf{narrative alignment}, where participants categorize each \textit{wh}-question based on the predominant narrative element it represents. This allows us to verify whether the human-identified narrative label matches the model-assigned one.
After these two initial assessments, participants are presented with the full MCQ, which now includes the \textit{wh}-question along with four answer choices: one correct answer and three distractors (without indicating which one is correct). At this stage, participants assess the (3) \textbf{clarity} of each MCQ option, identifying whether all answer choices are clearly written or if any present issues.
Next, we evaluate (4) \textbf{answerability}, where participants first indicate whether the answer to the \textit{wh}-question can be found in the text. If yes, they are instructed to select the correct answer from the four options or indicate that no correct answer is present.
Two additional metrics focus on the overall quality of the MCQ: (5) \textbf{plausability} of the incorrect answers, assessing whether the distractors are reasonable alternatives; and (6) \textbf{difficulty}, where participants rate the overall difficulty of the MCQ considering an 8-year-old child, the targeted age as explained in Section~\ref{sec:generation_rules}. Both plausibility and difficulty are rated using a 5-point Likert scale.
Finally, there is an (7) \textbf{observations} field where participants can optionally justify or comment on their ratings. This field proved invaluable for gaining a better understanding of the participants' assessments, which is why we have cited some of these observations throughout the results (Section~\ref{sec:expert_results}).

In Appendix~\ref{sec:expert_eval_form}, we show a detailed scheme of the actual form that has been shared with all participants.

\subsection{Results} \label{sec:expert_results}

This section presents the human evaluation results of generated MCQs, including well-formedness, narrative alignment, options clarity, answerability, plausibility, and difficulty.
Except for plausibility and difficulty, the values shown result from majority voting; for example, a question is considered well-formed if at least 2 out of 3 participants deem it so.

In some results tables, we have included a ``Not Eval'' column indicating the number of questions that were not analyzed for some metrics. This occurs when questions have already been classified as problematic in previous metrics, preventing the propagation of issues from earlier analyses and allowing for an independent evaluation of each metric.

\subsubsection{Well-formedness} \label{sec:results_q_form}

Table~\ref{tab:q_form} presents the well-formedness results for the \textit{wh}-question component of generated and human-authored MCQs.
While human-authored \textit{wh}-questions have the highest proportion of well-formed cases (42/45, 93.3\%), the questions generated by the models perform comparably, with \texttt{Ptt5-v2+Gemma-2} achieving the lowest score at 39/45 (86.7\%).
Among the well-formed questions, a recurring event is the generation of questions in a different Portuguese variant (Brazilian Portuguese). For instance, \texttt{Gemma-2} produces 8/45 well-formed questions in Brazilian Portuguese.
Notably, \texttt{Ptt5-v2+Gemma-2} exhibits only one such case. This is likely because the first step has used \texttt{Ptt5-v2} model, which has been fine-tuned exclusively on the European Portuguese variant (recall Section~\ref{sec:method_details}), mitigating this issue.

Regarding the questions classified as not well-formed, one human-authored question was flagged with a grammatical issue, which resulted from a typo and can be considered an outlier. Additionally, two human-authored questions were flagged for semantic issues. Upon further analysis, we found that these cases involved overly generic \textit{wh}-questions (e.g., ``What does the story tell?''\footnote{The following examples have been translated from Portuguese to English.}), which are not necessarily incorrect but may lack specificity.  
In fact, this type of semantic issue was the most common cause of ill-formed questions generated by the models. \texttt{Ptt5-v2+Gemma-2} exhibited 6/45 cases in which participants reported that the questions were unclear (e.g., ``Where did the cat go?''), thereby ``raising doubts and preventing the student from reaching the answer.''. \texttt{GPT-4o} and \texttt{Gemma-2}, with fewer occurrences (3/45 each), were also flagged for semantic issues due to nonsensical questions that did not align with actual events in the story (e.g., ``Why did Mom get angry with the eldest daughter?'', where, as noted by a participant, ``under no circumstances was the mother truly angry with her eldest daughter.'').

\begin{table}[h]
\caption{Human evaluation of well-formedness for the \textit{wh}-question component of generated and human-authored MCQs.}\label{tab:q_form}
\begin{tabular*}{\textwidth}{@{\extracolsep{\fill}}lccccccc}
\toprule
Provenance  & \# Eval  & \multicolumn{2}{c}{\# Well-Formed}  & \multicolumn{4}{c}{\# Not Well-Formed} \\  
\cmidrule(lr){3-4} \cmidrule(lr){5-8}  
            &         & WF  & WF (BR-Var)  & Ortho.  & Gram.  & Sem.  & Multi.  \\  
\midrule
Human                  & 45  & 42  & 0  & 0  & 1  & 2  & 0 \\
\texttt{GPT-4o}        & 45  & 40  & 2  & 0  & 0  & 3  & 0 \\
\texttt{Gemma-2}       & 45  & 33  & 8  & 0  & 1  & 3  & 0 \\
\texttt{Ptt5-v2+Gemma-2} & 45  & 38  & 1  & 0  & 0  & 6  & 0 \\
\botrule
\end{tabular*}
\footnotetext{WF: Well-formed, WF (BR-Var): Well-formed but in Brazilian Portuguese, Ortho: Orthographic errors, Gram: Grammatical errors, Sem: Semantic errors, Multi: Multiple errors.}
\end{table}

\subsubsection{Narrative Alignment} \label{sec:q_nar}

Table~\ref{tab:q_nar} presents the narrative alignment results for the \textit{wh}-question component of generated MCQs.
Overall, participants identified a narrative element that is aligned with the one pre-associated with the question. Only 2/39 cases, both from \texttt{Ptt5-v2+Gemma-2}, were identified not identified.
Upon further analysis, we found that these discrepancies stemmed from reasonable ambiguity regarding the predominant narrative element. For example, in the question ``Why was the girl so stunned?'', \textit{causal} narrative element was provided as input during generation, and thus remained associated with the resulting question. However, participants labeled it as \textit{feeling}. This divergence is understandable, as the question, while focusing on a causal event, also conveys aspects of the character’s emotions.
Similarly, in ``How did Mr. Pascoal feel when he found happiness?'', the \textit{feeling} narrative element was given as input, but participants categorized the question as referring to an \textit{action} event.

\begin{table}[h]
    \caption{Human evaluation of narrative alignment for the \textit{wh}-question component of generated MCQs.}
    \label{tab:q_nar}
    \centering
    \begin{tabular*}{\textwidth}{@{\extracolsep\fill}lccc c}
        \toprule
        Provenance\footnotemark[1] & \# Eval & Aligned & Not Aligned & \# Not Eval\footnotemark[2] \\
        \midrule
        \texttt{GPT-4o} & 42 & 42 {\footnotesize (100\%)} & 0 & 3 \\
        \texttt{Gemma-2} & 41 & 41 {\footnotesize (100\%)} & 0 & 4 \\
        \texttt{Ptt5-v2+Gemma-2} & 39 & 37 {\footnotesize (94.9\%)} & 2 & 6 \\
        \botrule
    \end{tabular*}
    \footnotetext[1]{Human-authored MCQs are not analyzed here, as they do not contain pre-annotated labels for narrative elements.}
    \footnotetext[2]{Number of MCQs not included in this analysis as the corresponding \textit{wh}-questions are classified as not well-formed (Table~\ref{tab:q_form}).}
\end{table}

\subsubsection{Options Clarity} \label{sec:mcq_form}

Table~\ref{tab:mcq_form} presents the clarity results of both generated and human-authored options for MCQs.
For human-authored MCQs, there are 2/42 flagged cases where participants indicated the need to reword the options for better clarity and understanding.
Regarding the models, multiple cases of problematic issues were flagged for the MCQ options. A common issue highlighted in the participants' comments is poor verb conjugation in the options, specifically options formulated with a verb tense that differs from the expected one or that do not agree with the subject of the question.
Another issue concerns whether the subject of option sentences should be explicitly stated rather than omitted. For example, in response to the question, ``What did the pink elephants do every day?'', the correct answer, ``Danced round and round in the moonlight of three moons.” could instead begin with “The elephants danced...'' for greater clarity.
Experts had differing views on whether declarative options should explicitly include the subject, which can enhance clarity but may also lead to redundancy.
Finally, the remaining issues relate to options that require rewording to better align with the text or to provide more specific information.

\begin{table}[h]
    \caption{Human evaluation of clarity for both generated and human-authored options for MCQs.}
    \label{tab:mcq_form}%
    \begin{tabular*}{\textwidth}{@{\extracolsep{\fill}}lccccc@{}}
        \toprule
        Provenance & \# Eval & Clear & Not Clear & \# Not Eval\footnotemark[1] \\ 
        \midrule
        \texttt{Human} & 42 & 40 {\footnotesize (95.2\%)} & 2 & 3 \\
        \texttt{GPT-4o} & 42 & 38 {\footnotesize (90.5\%)} & 4 & 3 \\
        \texttt{Gemma-2} & 41 & 35 {\footnotesize (85.4\%)} & 6 & 4 \\
        \texttt{Ptt5-v2+Gemma-2} & 39 & 37 {\footnotesize (94.9\%)} & 2 & 6 \\
        \botrule
    \end{tabular*}
    \footnotetext[1]{Number of MCQs not included in this analysis as the corresponding \textit{wh}-questions are classified as not well-formed (Table~\ref{tab:q_form}).}
\end{table}

\subsubsection{Answerability} \label{sec:mcq_answer}
Table~\ref{tab:mcq_answer} shows the human evaluation of answerability for generated and human-authored MCQs. 
This metric revealed various issues across all types of MCQs, including those originating from human authors. The most common problem is \textit{NCA-A}, where participants indicate that there is an answer to the \textit{wh}-question in the text, but none of the MCQ options match it. Specifically, one participant reported that ``The correct option is not completely in line with what is observed in the text.'' There are also significant \textit{NCA} cases, where participants stated that the correct answer does not exist in the text, making all options incorrect. As one participant noted, ``The correct answer is not explicit in the text.'', highlighting problems of explicitness. Additionally, \texttt{Ptt5-v2+Gemma-2} presented undesirable MCQs containing multiple valid correct answers or cases where the human choice does not match the labeled correct answer, indicating ambiguity in the generated MCQs.

\begin{table}[!ht]
\caption{Human evaluation of answerability for generated and human-authored MCQs.}\label{tab:mcq_answer}%
\begin{tabular*}{\textwidth}{@{\extracolsep{\fill}}lcccccccc}
\toprule
Provenance  & \# Eval  & \multicolumn{1}{c}{Ans.}  & \multicolumn{5}{c}{Not Ans.\footnotemark[1]} & \# Not Eval\footnotemark[2] \\  
\cmidrule(lr){4-8}  
            &         &   & NCA  & NCA-A  & MVA & MCA & MUL & \\  
\midrule
\texttt{Human}             & 40  & 33 {\footnotesize (82.5\%)}  & 1  & 5  & 0  & 0  & 1  & 5 \\
\texttt{GPT-4o}           & 38  & 33 {\footnotesize (86.8\%)}  & 1  & 4  & 0  & 0  & 0  & 7 \\
\texttt{Gemma-2}          & 35  & 33 {\footnotesize (94.3\%)}  & 1  & 1  & 0  & 0  & 0 & 10 \\
\texttt{Ptt5+Gemma-2}     & 37  & 25 {\footnotesize (67.6\%)}  & 5  & 0  & 2  & 3  & 2  & 8 \\
\botrule
\end{tabular*}
\footnotetext[1]{The sub-columns under ``Not Answerable'' represent:\\ 
NCA: No Correct Answer – The correct answer does not exist in the text, so none of the options are correct.\\  
NCA-A: No Correct Answer (but Answer exists) – The correct answer exists in the text, but none of the options match it.\\  
MVA: Multiple Valid Answers – More than one option is the correct answer.\\  
MCA: Mismatched Correct Answer – Human choice does not match the labeled correct answer.\\  
MUL: Multiple Problems – More than one issue affects the answerability.}
\footnotetext[2]{Number of MCQs not included in this analysis as the corresponding \textit{wh}-questions are classified as not well-formed (Table~\ref{tab:q_form}), or the MCQ options are classified as unclear (Table~\ref{tab:mcq_form}).}
\end{table}

\subsubsection{Distractor Plausability} \label{sec:plausability}

\begin{figure}[!ht]
    \centering
    \includegraphics[width=\textwidth]{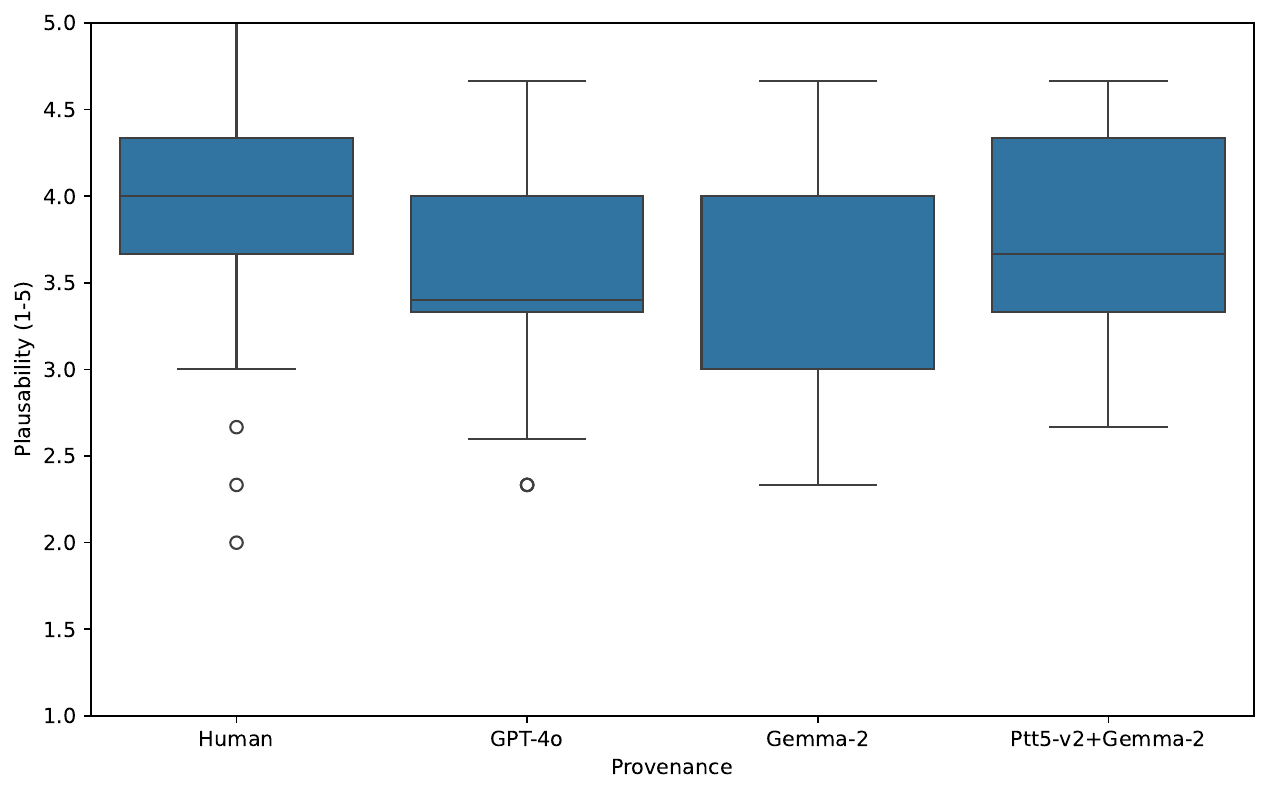}
    \caption{Expert-perceived MCQ options plausability by provenance.}
    \label{fig:plausability}
\end{figure}

Figure~\ref{fig:plausability} shows the distribution values of expert-perceived MCQ distractor plausibility by provenance.
Distractors in human-authored MCQs are perceived as the most plausible (mean: 3.93, median: 4.0), while \texttt{Gemma-2} distractors are perceived as the least plausible (mean: 3.61, median: 4.0).
We perform a Kruskal-Wallis test, which is a statistical test\footnote{One-way ANOVA was not applied because, according to the normality test using Shapiro-Wilk, the values are not normally distributed. The Kruskal-Wallis test is more appropriate for this data as it does not assume normality.} used to compare independent groups, to determine if there are statistically significant differences among the various MCQ samples based on provenance. We obtained the following result: H=5.585, \( p\text{-value} = 0.134 \). As $p > 0.05$, we conclude that there are no statistically significant differences in distractor plausibility among the different provenances of MCQs.

\subsubsection{Difficulty} \label{sec:results_difficulty_human}

\begin{figure}[!ht]
    \centering
    \includegraphics[width=\textwidth]{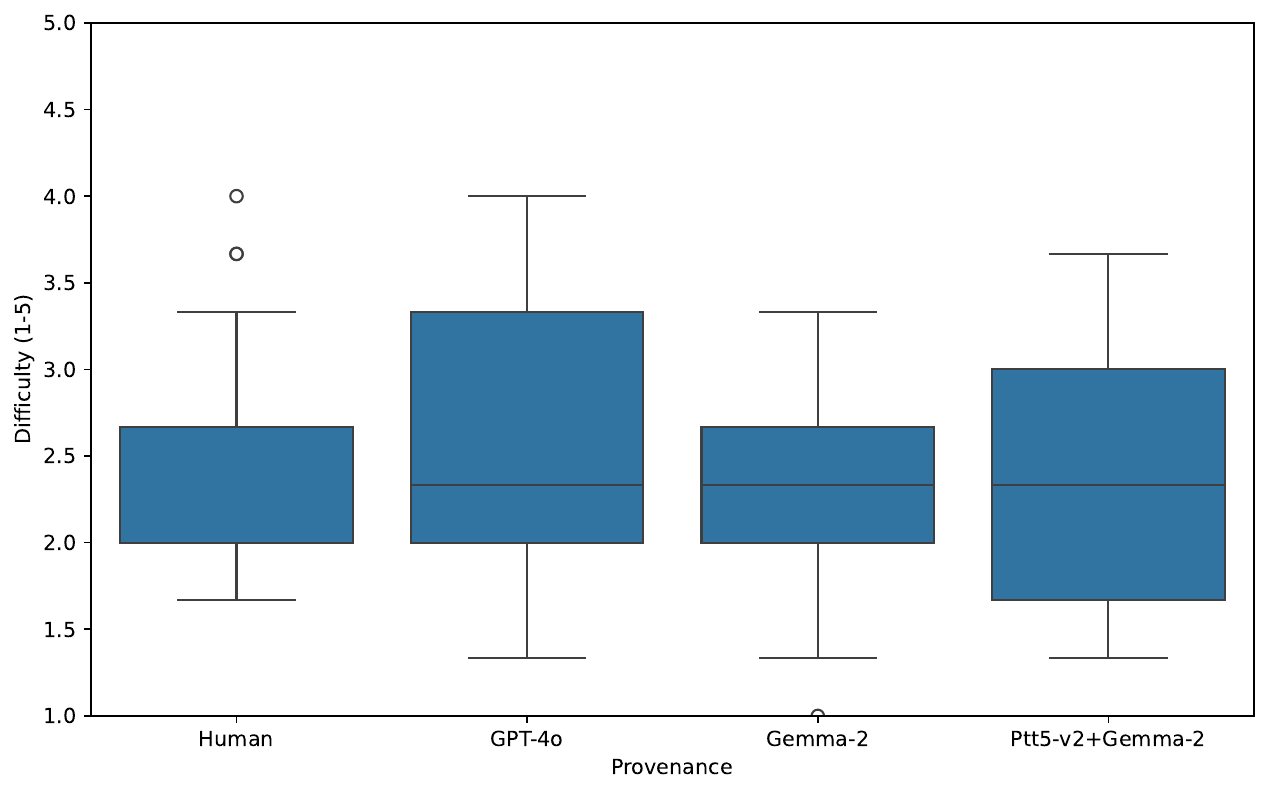}
    \caption{Expert-perceived MCQ difficulty by provenance.}
    \label{fig:difficulty_human}
\end{figure}

Figure~\ref{fig:difficulty_human} shows the distribution values of expert-perceived MCQ difficulty by provenance.
As in plausibility, human-authored MCQs are perceived as the most difficult (mean: 2.49, median: 2.67), while \texttt{Gemma-2} MCQs are perceived as the least difficult (mean: 2.27, median: 2.33).
The Kruskal-Wallis test outputs the following result: \( H = 1.433 \), \( p\text{-value} = 0.698 \). As \( p > 0.05 \), we conclude that there are no statistically significant differences among the different provenances of MCQs regarding difficulty.
In general, these results indicate that, according to human perception, most of the MCQs are easy (mean: \( < 2.5 \)) for an 8-year-old student.

\subsection{Discussion of Expert Review Results}

We revisit our first research question (RQ1): \textit{How acceptable are MCQs generated by current generative models?}

In terms of well-formedness, the proportion of well-formed \textit{wh}-questions ranges from 86.7\% (\texttt{Ptt5-v2+Gemma-2}) to 93.3\% (Human and \texttt{GPT-4o}).
Therefore, we found that most \textit{wh}-questions generated by current models are acceptable according to experts and align with the quality of human-authored questions.
We also discovered that among the well-formed questions, a recurring issue with current models is the generation of \textit{wh}-questions in a different language variant (i.e., Brazilian Portuguese in our study case). This phenomenon has already been documented in the literature~\citep{leite_2024_fairy_pt}, and it remains a limitation. Since these models are trained on vast amounts of data across different language variants, they do not always strictly follow prompt instructions, even when explicitly asked to generate text in a specific variant, as noted in our prompt (see Section~\ref{sec:final_prompt}).
One solution to this is to use fine-tuning on data exclusive to the target language variant (as has been done with \texttt{Ptt5-v2+Gemma-2}). However, models trained in this way may have limited generalization capabilities, as they are optimized primarily for the specific characteristics and coverage of the fine-tuning data.
Among the ill-formed questions, we found that the most common issues are semantic errors, including overly generic, unclear, or nonsensical \textit{wh}-questions. This is true for both human-authored and generated questions, demonstrating that they share the same types of problems.

Regarding narrative alignment, we found that most generated questions are perceived as aligned with the target narrative elements. This alignment is ideal in \texttt{GPT-4o} and \texttt{Gemma-2}, demonstrating the effectiveness of narrative control through a zero-shot prompting strategy.
While 5\% of the questions generated by \texttt{Ptt5-v2+Gemma-2} were perceived as misaligned, we found that these questions can be seen as containing two plausible narrative elements.
Therefore, our results indicate that controlling narrative elements through zero-shot prompting or fine-tuning is effective.

Regarding the clarity of MCQ options, we found that the acceptance rate of \texttt{GPT-4o} and \texttt{Gemma-2} MCQs (90.5\% and 85.4\%, respectively) is slightly lower than that of human-authored MCQs (95.2\%). 
This suggests that current models may be more prone to generating unclear options. According to the experts, this issue arises primarily due to improper subject-verb agreement, incorrect verb tenses, or phrasings that require rewording for better clarity and alignment with the source text.
A common suggestion from some experts has been to include (rather than omit) the subject in MCQ options. After further discussion, we observed differing perspectives on whether declarative options should explicitly begin with the subject embedded in the sentence. On the one hand, this can lead to lengthy, repetitive sentences. On the other hand, simplifying the options by omitting the subject (when it is implied) can improve readability. Indeed, subject omission varies depending on contextual factors~\citep{haegeman_2007_subject}.
Our prompt did not include a specific instruction to include or omit the subject in the options, as we believe that this type of conditioning can vary depending on the context.

Regarding the answerability of MCQ options, we found that the acceptance rate of \texttt{GPT-4o} and \texttt{Gemma-2} MCQs (86.9\% and 94.3\%, respectively) is slightly higher than that of human-authored MCQs (82.5\%).
This contrasts with the findings for option clarity and suggests that these models are more capable of generating answerable MCQs.
The most commonly reported issue has been the situation where none of the options is the correct answer, despite the answer being recognized in the source text by the experts. This problem primarily affects human-authored and \texttt{GPT-4o} MCQs, revealing again that they share the same type of problems.
Additionally, the issue of the text source not including the answer (meaning none of the options could be correct) has been reported across all MCQ provenances, particularly for \texttt{Ptt5-v2+Gemma-2}, which shows a relatively low acceptance rate for this metric (67.6\%).
We believe this is due to the nature of the two-step method (recall Section~\ref{sec:method_two_step}): the first step generates the \textit{wh}-question, while the second generates the remaining components. Because these are two different modules, the first step may impact the second regarding the agreement between the question and the answer. This indicates that using a two-step generation process for creating MCQs may not be effective.
Overall, the problems reported in answerability are related to the generally recognized issues in determining the correct choice for an MCQ, which necessitates thorough validation~\citep{haladyna_2004_mcq_validation} during the creation of MCQs prior to administration to test-takers.

Regarding the plausibility of distractors, experts tended to rate human-authored MCQs slightly higher, but no statistically significant differences were found between different MCQ provenances. This suggests that the performance of current models is broadly comparable to human-authored MCQs in terms of distractor plausibility.
Similarly, in terms of difficulty, no significant differences were observed between MCQ provenances, highlighting that most MCQs (whether generated or not) are generally perceived as easy (mean: \( < 2.5 \)).
%
In Section~\ref{sec:perceived_difficult}, we further analyze the correlation between expert perceptions and model-annotated difficulty values.

In general, experts indicate that MCQs generated by LLMs (\texttt{GPT-4o} and \texttt{Gemma-2}) exhibit the same types of flaws as human-authored MCQs, although the distribution of these flaws differs.
Interestingly, both LLM-generated and human-authored MCQs ultimately reached the same overall acceptance rate ($\frac{33}{45}$, 73.3\%) --- obtained from the answerability metric (Table~\ref{tab:mcq_answer}). This suggests that their quality is comparable. However, \texttt{Ptt5-v2+Gemma-2}, which employs a two-step approach, demonstrated significant issues with answerability, resulting in an overall acceptance rate ($\frac{25}{45}$, 55.6\%), which is substantially lower than that of human-authored MCQs.

\section{Evaluation: Psychometric Properties} \label{sec:eval_students}

This section describes the evaluation of MCQs using psychometric properties derived from student answers. It covers participant characterization, data selection, evaluation procedure, classical test theory, and the results obtained.

\subsection{Participants}

We engaged with \textit{Agrupamento de Escolas de S\~ao Louren\c{c}o}\footnote{\url{https://agrupamentoslourenco.org/}}, a group of schools located in the Valongo region (district of Porto), where some teachers agreed to collaborate in this evaluation by employing MCQs with their students. A total of 14 classes participated (284 students).

\subsection{Data Collection}

The MCQs presented to students are those previously evaluated by experts, as described in Section~\ref{sec:eval_expert}. Since it is not pedagogically appropriate to present flawed MCQs to students, we have only used those that experts deemed free of issues: MCQs that are well-formed, answerable, and with options that are clearly written.
We did not apply any filtering based on plausibility or difficulty ratings, as we believe it is important to have a representative sample of MCQs with varying ratings for these metrics.
For MCQs that were considered well-formed but written in the Brazilian Portuguese variant, we manually adapted them to the European Portuguese.

\subsection{Evaluation Procedure}

The evaluation protocol can be summarized as follows:

\begin{enumerate}
    \item \textbf{MCQ Filtering}: We used only validated MCQs (well-formed, clear options, and answerable). A total of 124 MCQs were selected, resulting from the answerability analysis (recall Table~\ref{tab:mcq_answer}). Specifically, this includes 33 human-authored, 33 from \texttt{GPT-4o}, 33 from \texttt{Gemma-2}, and 25 from \texttt{Ptt5-v2+Gemma-2}.
    \item \textbf{Building Paper Sheets from Forms}: Given the infeasibility of using Google Forms with children in all schools, we converted each form into a paper sheet.
    Each sheet contains an alphanumeric code for tracking each student's answers. No personal data was collected.
    \item \textbf{Sheet Distribution}: The sheets were distributed among the classes, with each class answering to two sheets containing texts appropriate for their grade level. The distribution of paper sheets per class can be observed in Table~\ref{tab:class_dist}.
    \item \textbf{Student Responses}: Each sheet was completed in a classroom environment under teacher supervision to ensure individual responses. Of the 124 MCQs used in this evaluation, we obtained an average of 45 individual responses per MCQ. The MCQ with the fewest responses has 39, while the one with the most has 55.
    \item \textbf{Sheet Collection}: Once completed, the sheets were collected, and student responses were aggregated for psychometric analysis.
\end{enumerate}

This process took place between February 3 and February 21, 2025.

\begin{table}[h]
    \caption{Overview of classes, number of students and sheets assignment.}
    \label{tab:class_dist}
    \centering
    \begin{tabular*}{\textwidth}{@{\extracolsep\fill}lcccc}
        \toprule
        Class & Year & \# Students (\footnotemark[1]) & Form ID (1st Sheet) & Form ID (2nd Sheet) \\
        \midrule
        C1 & 2 & 19 (1)  & 1 & 2 \\
        C2 & 3 & 24 (1)  & 1 & 2 \\
        C3 & 3 & 23 (0)  & 3 & 4 \\
        C4 & 3 & 18 (0)  & 3 & 4 \\
        C5 & 3 & 18 (1)  & 5 & 6 \\
        C6 & 3 & 17 (3)  & 5 & 6 \\
        C7 & 3 & 20 (1)  & 5 & 6 \\
        C8 & 3 & 20 (1)  & 7 & 8 \\
        C9 & 3 & 24 (0)  & 7 & 8 \\
        C10 & 4 & 18 (0) & 9 & 10 \\
        C11 & 4 & 19 (3) & 9 & 10 \\
        C12 & 4 & 20 (4) & 9 & 10 \\
        C13 & 4 & 22 (0) & 11 & 12 \\
        C14 & 4 & 22 (0) & 11 & 12 \\
        \midrule
        Total &  & 284 (15) &  &  \\
        \botrule
    \end{tabular*}
    \footnotetext[1]{Values in parentheses represent the number of students with special educational needs.}
\end{table}

\subsection{Classical Test Theory}

We use Classical Test Theory (CTT)~\citep{crocker_86_ctt} to analyze student responses. CTT serves as a foundational measurement tool in psychometrics for estimating the quality of test items. Specifically, we focus on estimating two key parameters for each item (MCQ): difficulty and discrimination.

\textbf{\(P\)}\footnote{We have used the capital \(P\) to avoid confusion with the \(p\)-value of a significance test.} is the proportion of students who answer correctly:
\[
P = \frac{X}{N}
\]
where \(X\) represents the number of students who answered the item correctly, and \(N\) is the total number of students who attempted the item.
While this is sometimes referred to as a ``difficulty index'' in the CTT literature, it more accurately reflects easiness. For clarity and consistency, we report MCQ difficulty in our results as 1 - \(P\), which ranges from 0 to 1, with higher values indicating more difficult items.

\textbf{\(D\)} (also known as \textbf{discrimination}) is obtained by calculating the difference between the proportions of high-performing and low-performing students who answer the item correctly:
\[
D = {\textit{high}} - {\textit{low}}
\]
where \({\textit{high}}\)  is the proportion of correct responses from the top \(27\%\) of performers, and \({\textit{low}}\) is the proportion from the bottom \(27\%\) of performers. A positive discrimination value indicates that the item effectively differentiates between high and low performers, while a negative value suggests that the item may not be effective in distinguishing between these groups.

\subsection{Results: MCQ Difficulty and Discrimination} \label{sec:results_pd}

\begin{figure}[!ht]
    \centering
    \includegraphics[width=\textwidth]{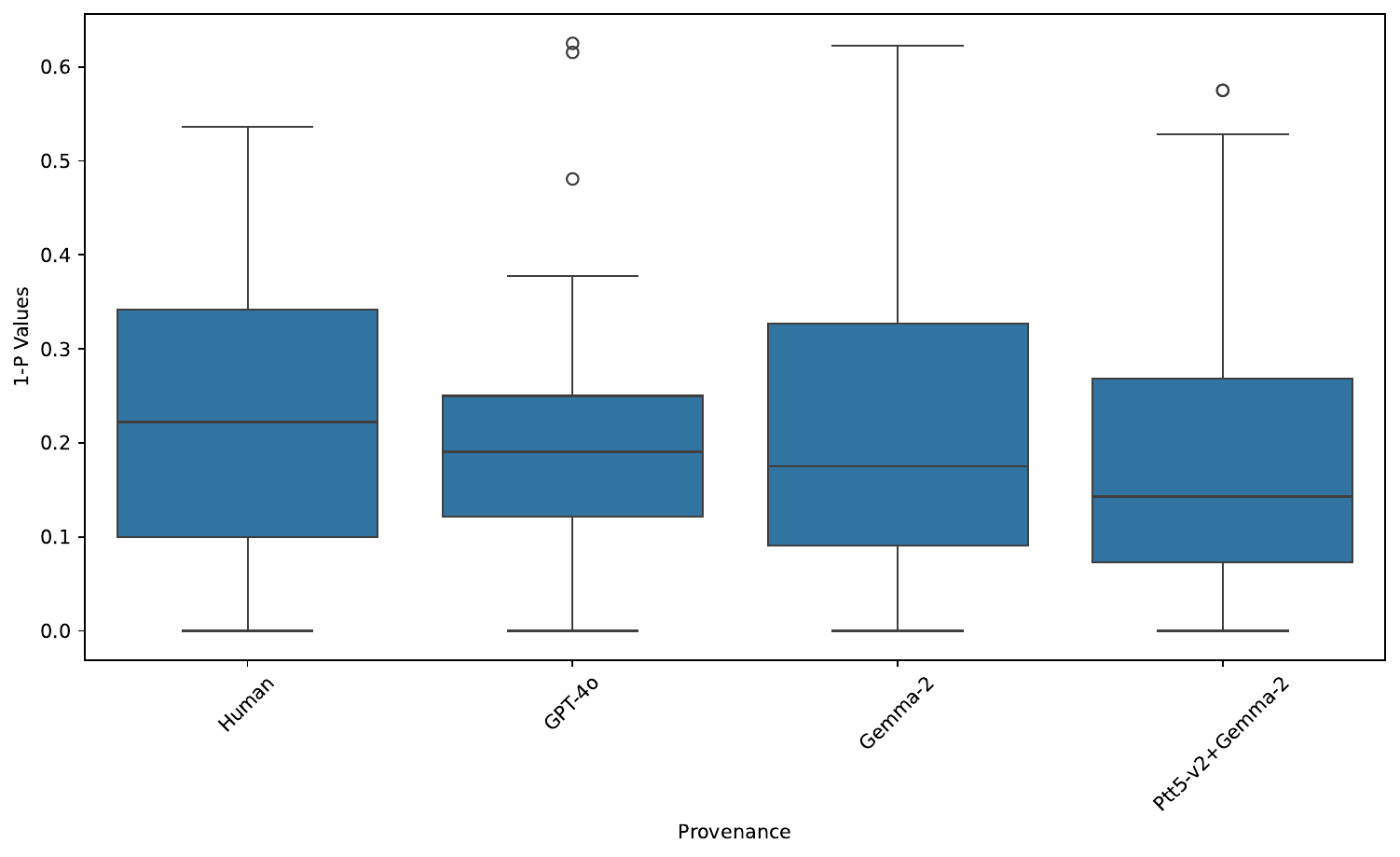}
    \caption{Distribution of estimated \(1-P\) (difficulty) values by MCQs provenance.}
    \label{fig:p_values}
\end{figure}

Figure~\ref{fig:p_values} shows the distribution of estimated \(1 - P\)\footnote{We invert the values (\(1 - P\)) so that higher scores represent higher difficulty.} (difficulty) values from student responses.  
Human-authored MCQs are perceived as the most difficult (\(\text{mean} = 0.232\)), while \texttt{Ptt5-v2+Gemma-2} MCQs are perceived as the least difficult (\(\text{mean} = 0.199\)).  
The estimated difficulty is generally low for all MCQs, since the mean is \(\leq 0.232\) for all MCQ provenances.
Overall, there are no statistically significant differences between the groups, as indicated by the Kruskal-Wallis test (\( H = 1.249 \), \( p\text{-value} = 0.741 \)), with \( p > 0.05 \).

\begin{figure}[!ht]
    \centering
    \includegraphics[width=\textwidth]{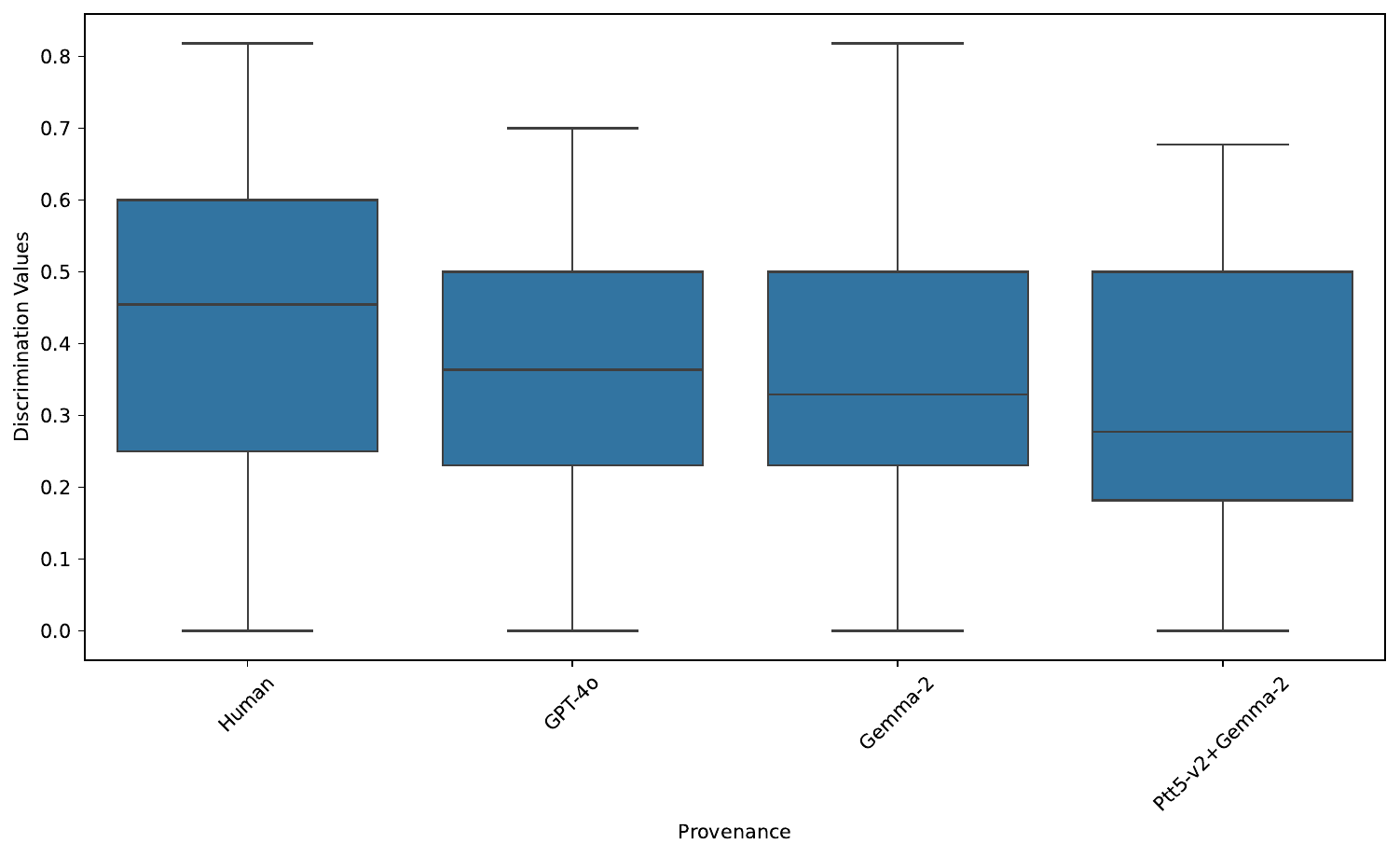}
    \caption{Distribution of estimated \(D\) (discrimination) values by MCQs provenance.}
    \label{fig:discrimination_values}
\end{figure}

Figure~\ref{fig:discrimination_values} shows the distribution of estimated \(D\) (discrimination) values.
Again, human-authored MCQs are perceived as the most discriminative (\(\text{mean} = 0.414\)), while \texttt{Ptt5-v2+Gemma-2} MCQs are perceived as the least discriminative (\(\text{mean} = 0.336\)).  
The MCQs are estimated to have reasonable discriminative values, since the mean is \( \geq 0.336\) considering all MCQ provenances.  
Overall, there are no statistically significant differences between the groups, as indicated by the one-way ANOVA test (\( H = 0.854 \), \( p > 0.05 \)). 

\subsection{Results: Reliability of MCQ Options}

The quality of MCQs can also be assessed through the reliability of their options, including both the correct answer and the distractors, based on student responses. First, we examine which MCQs have more distractors selected by students. 
Table~\ref{tab:3_dist_used} displays the number of MCQs by provenance where each distractor has been selected at least once.

\begin{table}[h]
    \caption{Number of MCQs by provenance where each distractor has been selected at least once.}
    \label{tab:3_dist_used}%
    \begin{tabular*}{\textwidth}{@{\extracolsep{\fill}}lccc@{}}
        \toprule
        Provenance & \# Eval & 3-Dist-Used\footnotemark[1] (Yes) & 3-Dist-Used\footnotemark[1] (No) \\ 
        \midrule
        \texttt{Human} & 33 & 19 {\footnotesize (57.6\%)}  & 14 \\
        \texttt{GPT-4o} & 33 & 15 {\footnotesize (45.5\%)} & 18 \\
        \texttt{Gemma-2} & 33 & 17 {\footnotesize (51.5\%)} & 16 \\
        \texttt{Ptt5-v2+Gemma-2} & 25 & 9 {\footnotesize (36.0\%)} & 16 \\
        \botrule
    \end{tabular*}
    \footnotetext[1]{The 3 distractors are used at least once.}
\end{table}

We found that human-authored MCQs have the highest proportion of student-selected distractors (57.6\%) and \texttt{Ptt5-v2+Gemma-2} has the least (36.0\%).
These results show that students engaged more with distractors present in human-authored MCQs.

We also attempt to identify MCQs with problematic options by applying the three principles adopted by \citet{martinkova_2018_shinyitemanalysis} and \citet{lin_2024_psyco} (3-rule test):
\begin{enumerate}
    \item The proportion of high-ability respondents selecting the correct option should be greater than the proportion of high-ability respondents selecting any of the distractors for that MCQ.
    \item The proportion of respondents selecting the correct option should monotonically increase from low- to high-ability respondents.
    \item The proportion of respondents selecting any distractor should monotonically decrease from low- to high-ability respondents.
\end{enumerate}
%
To check the three principles, we first divide the student sample into three groups: low-, medium-, and high-ability. 
Here, ability is estimated based on each student's overall performance across the set of MCQs they answered (represented by $\theta$). Students below the first quartile (Q1) are classified as low-ability, those between Q1 and Q3 as medium-ability, and those above Q3 as high-ability.
This three-group division follows prior work~\citep{lin_2024_psyco} and offers a practical balance between interpretability and group size --- finer subdivisions would result in smaller groups, potentially limiting the robustness of the analysis.

\begin{figure}[!ht]  
    \centering
    \begin{subfigure}{0.45\textwidth}
        \centering
        \includegraphics[width=\linewidth]{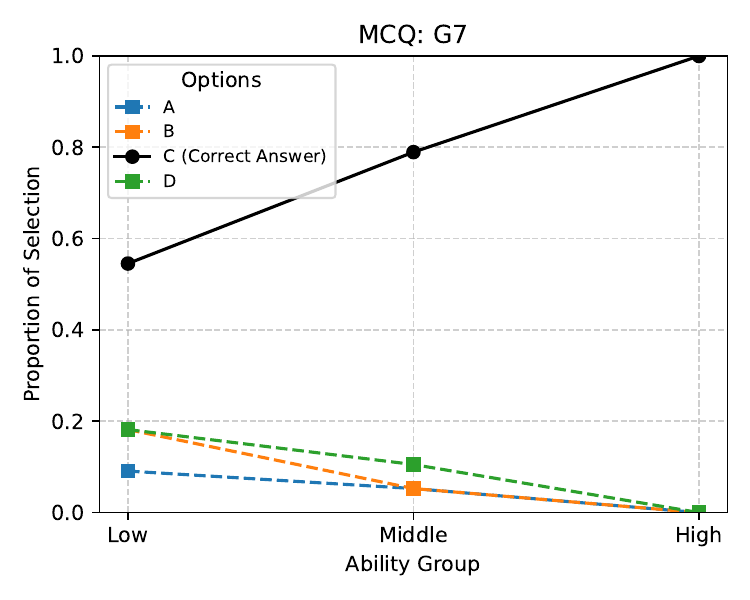}
        \caption{All 3 rules pass.}
    \end{subfigure}
    \hfill
    \begin{subfigure}{0.45\textwidth}
        \centering
        \includegraphics[width=\linewidth]{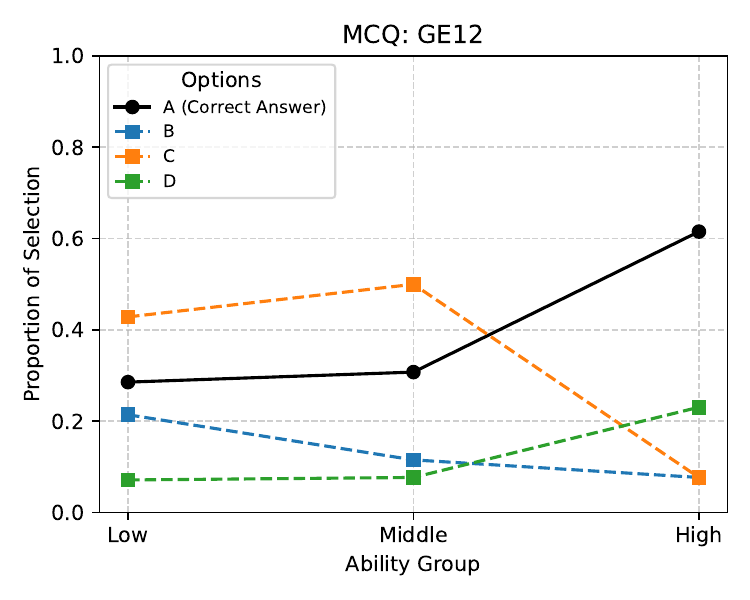}
        \caption{Rule 3 fails.}
    \end{subfigure}

    \begin{subfigure}{0.45\textwidth}
        \centering
        \includegraphics[width=\linewidth]{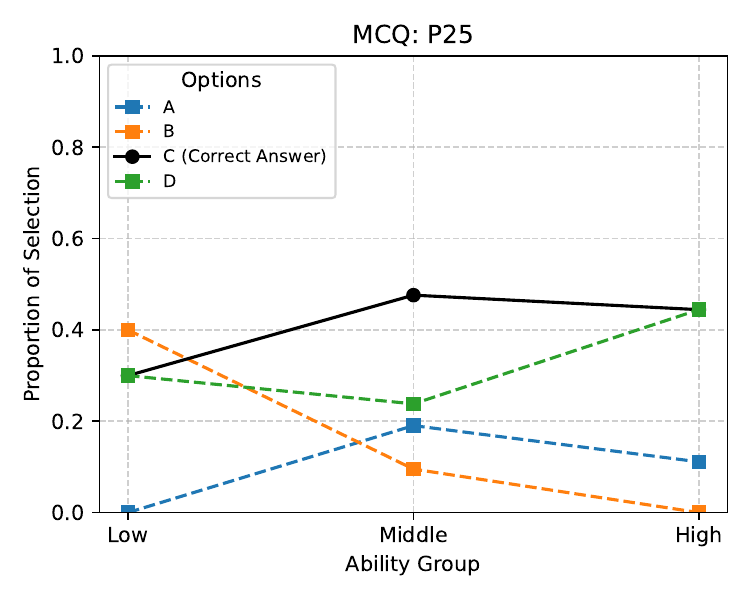}
        \caption{All 3 rules fail.}
    \end{subfigure}

    \caption{Proportion of MCQs options selection by ability group (some examples).}
    \label{fig:examples_3rules_test}
\end{figure}

Figure~\ref{fig:examples_3rules_test} illustrates MCQs that either follow or violate these principles. For instance, in case (b), rule 3 fails because the proportion of respondents selecting distractor D increases (instead of decreasing) from the middle- to high-ability group.

%

Table~\ref{tab:rule_compliance} presents the overall results for rule compliance across different MCQ provenances. Here, we consider only MCQs where all distractors have been selected at least once (reported in Table~\ref{tab:3_dist_used}), ensuring a fair assessment for rule 3.
\texttt{Human} MCQs have the highest proportion of MCQs that pass all three rules (31.6\%), followed by \texttt{GPT-4o} (26.7\%), \texttt{Gemma-2} (23.5\%), and \texttt{Ptt5-v2+Gemma-2} (22.2\%).  
These results indicate that the number of MCQs complying with all three rules is generally low.
This is primarily due to rule 3, which is the most frequently violated rule across all provenances. This is understandable, as a single respondent from a higher-ability group selecting a distractor that was not chosen by a lower-ability group is enough to break the rule. While informative, this rule is naturally sensitive to small deviations.

\begin{table}[!ht]
\caption{Number of MCQs that comply (and do not) with the 3-rule test. Each subcolumn under ``Failed at Least One Rule'' indicates the number of MCQs that did not pass the respective rule. Since a question can fail multiple rules, row sums may exceed \# Eval.}
\label{tab:rule_compliance}
\begin{tabular*}{\textwidth}{@{\extracolsep{\fill}}lcccccc}
\toprule
Provenance  & \# Eval  & \multicolumn{1}{c}{Passed All 3 Rules}  & \multicolumn{3}{c}{Failed at Least One Rule} \\  
\cmidrule(lr){4-6}  
       &         &   & Failed Rule 1  & Failed Rule 2  & Failed Rule 3  \\  
\midrule
\texttt{Human}  & 19  & 6 {\footnotesize (31.6\%)}  & 0  & 0  & 13  \\  
\texttt{GPT-4o}  & 15  & 4 {\footnotesize (26.7\%)}  & 1  & 1  & 11  \\  
\texttt{Gemma-2}  & 17  & 4 {\footnotesize (23.5\%)}  & 0  & 0  & 13  \\  
\texttt{Ptt5-v2+Gemma-2}  & 9  & 2 {\footnotesize (22.2\%)}  & 1  & 1  & 7  \\  
\botrule
\end{tabular*}
\end{table}

\subsection{Discussion of Psychometric Properties Results} \label{sec:discussion_psycho}

We revisit our second research question (RQ2): \textit{How do the psychometric properties of generated MCQs compare to those of human-authored ones?}

In terms of estimated difficulty and discrimination values, we identify three key findings that align with experts. First, the overall difficulty is generally low for both human-authored and generated MCQs. Second, while there are no statistically significant differences between groups, human-authored MCQs are found to be the most difficult and discriminative, although by a small margin. Third, \texttt{Ptt5-v2+Gemma-2} MCQs are perceived as the least difficult and discriminative.

We also find that human-authored MCQs have more student-selected distractors and better adhere to option selection principles (3-rule test). This reinforces our findings: students engage more with human-authored distractors, suggesting higher plausibility due to their nuanced answer options.
Of particular note, MCQs from \texttt{Ptt5-v2+Gemma-2} exhibit the lowest proportion of student-selected distractors and the poorest adherence to option selection principles, confirming that the two-step method is shown to be the least unreliable for generating MCQs.

Overall, we interpret these findings in two ways. On one hand, human-authored MCQs remain a benchmark that generative models have not yet surpassed in terms of difficulty and discrimination. On the other hand, the performance of LLMs, such as \texttt{GPT-4o} and \texttt{Gemma-2}, is approaching human levels. Therefore, we conclude that these LLMs generate MCQs that are generally comparable to human-authored ones in terms of difficulty, discrimination, and reliability.

\section{Perceived Difficulty Characteristics of Generated MCQs} \label{sec:perceived_difficult}

The proposed methods (recall Section~\ref{sec:methods_mcq_gen}) aim to generate MCQs with varying levels of difficulty. In this section, we examine the perceived difficulty of generated MCQs across three perspectives:

\begin{itemize}
    \item \texttt{Experts-Dif}: Difficulty values (1–5) assigned by experts (recall Section~\ref{sec:results_difficulty_human}).
    \item \texttt{Students-Dif}: Difficulty values (0–1) estimated from student answers using psychometric properties (recall Section~\ref{sec:results_pd}).
    \item \texttt{\textit{ModelName}-Dif}: Difficulty values (0–100) assigned by generative models in the generation process (recall Section~\ref{sec:methods_mcq_gen}).
\end{itemize}

In Section~\ref{sec:difficulty_inter_perspective}, we examine how the different difficulty perspectives correlate with each other (inter-perspective analysis).
In Section~\ref{sec:difficulty_control_means}, we evaluate how well model-annotated difficulty values match MCQs' difficulty as perceived by experts and students.
In Section~\ref{sec:difficulty_out_perspective}, we analyze how various MCQ-level features (e.g., question length, semantic similarity) influence the perceived difficulty within each perspective.

\subsection{Inter-Perspective Correlation} \label{sec:difficulty_inter_perspective}

\begin{figure}[htbp]
    \centering
    \begin{subfigure}[b]{0.49\textwidth}
        \centering
        \includegraphics[width=\textwidth]{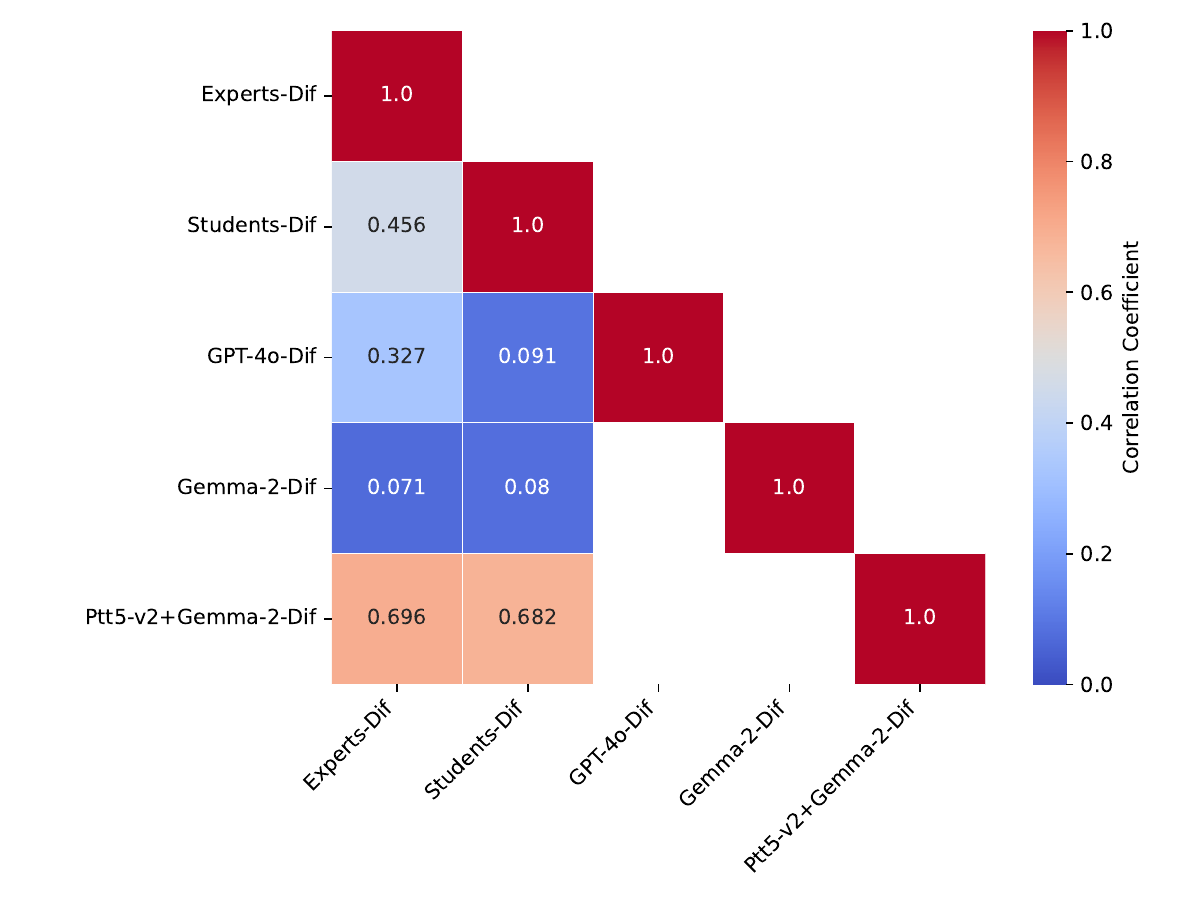}
        \caption{\texttt{GPT-4o-Dif} and \texttt{Gemma-2-Dif} values are predicted \textbf{in-generation}.}
        \label{fig:correlation_triangle_in}
    \end{subfigure}
    \hfill
    \begin{subfigure}[b]{0.49\textwidth}
        \centering
        \includegraphics[width=\textwidth]{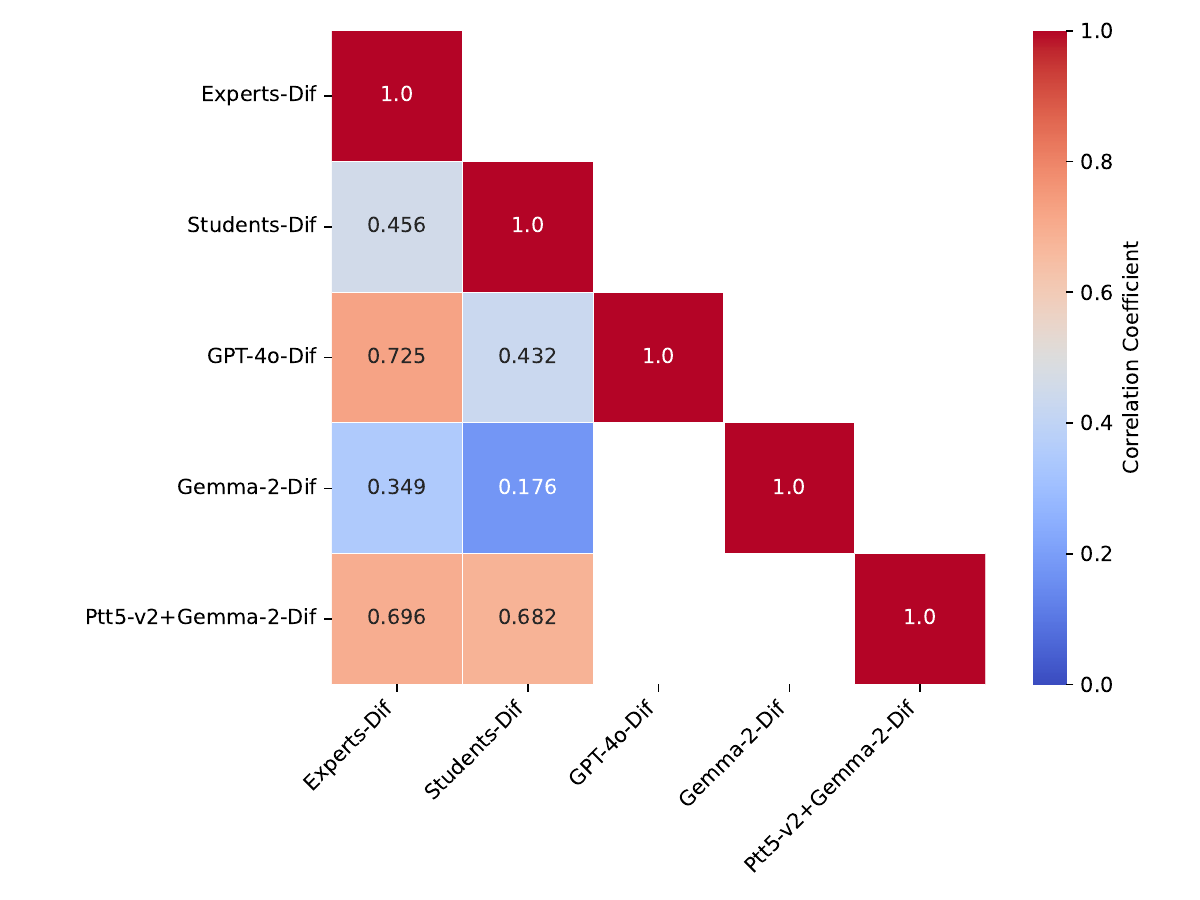}
        \caption{All \texttt{\textit{ModelName}-Dif} values are predicted \textbf{post-generation}.}
        \label{fig:correlation_triangle_post}
    \end{subfigure}
    \caption{Pearson correlation matrix between all \texttt{-Dif} perspectives. \texttt{\textit{ModelName}-Dif} instances have no inter-correlation since each model only annotated its own MCQs.}
    \label{fig:pearson_corr}
\end{figure}

The inter-perspective Pearson correlation values are presented in Figure~\ref{fig:pearson_corr}.
Figure~\ref{fig:correlation_triangle_in} shows the correlation values obtained using our original method. In this setup, \texttt{GPT-4o-Dif} and \texttt{Gemma-2-Dif} values are predicted via the \textbf{in}-generation process, whereas \texttt{Ptt5-v2-Gemma-2-Dif} values are predicted \textbf{post}-generation (according to the two-step method).
To investigate the impact of difficulty prediction timing, Figure~\ref{fig:correlation_triangle_post} presents correlation results when all difficulty values are predicted post-generation\footnote{That is, we conduct an additional experiment using a prompt that instructs \texttt{GPT-4o} and \texttt{Gemma-2} to predict difficulty based on the already generated full MCQ.}.

As a first observation, when difficulty values are predicted post-generation, the overall correlation values improve compared to in-generation prediction. This suggests that predicting difficulty after the full MCQ has been generated leads to better alignment with both expert perception and student performance.

As a second observation, we find a noticeable (though weak) correlation between \texttt{Experts-Dif} and \texttt{Students-Dif}. This implies that expert and student perceptions of difficulty are only slightly aligned.
To assess which of these better aligns with model-annotated difficulty values, we observe that all \texttt{\textit{ModelName}-Dif} values correlate more strongly with \texttt{Experts-Dif} than with \texttt{Students-Dif} --- in the post-generation setup\footnote{While this trend holds for all models, the correlation difference is minimal for \texttt{Ptt5-v2+Gemma-2-Dif}, with values of 0.696 (Experts) and 0.682 (Students).}.
This indicates that model-annotated difficulty values are more closely aligned with expert perception than with student performance.
For example, we observe strong correlations between \texttt{Experts-Dif} and \texttt{GPT-4o-Dif}, as well as between \texttt{Experts-Dif} and \texttt{Ptt5-v2-Gemma-2-Dif}.

\subsection{Model-Annotated Difficulty vs. Human Perception} \label{sec:difficulty_control_means}

Here we investigate whether MCQs with higher \texttt{\textit{ModelName}-Dif} values correspond to higher difficulty as perceived by \texttt{Experts-Dif} and \texttt{Students-Dif}, indicating how well model-assigned difficulty values align with human judgments.
To do this, for each model, the generated MCQs are divided into two groups: one with below-median (\textbf{Low}) and one with above-median (\textbf{High}) difficulty values.
Table~\ref{tab:difficulty_control_means} presents the \texttt{Experts-Dif} and \texttt{Students-Dif} mean difficulty values for each group.

\begin{table}[ht]
    \caption{Mean difficulty values from expert and student perspectives for MCQs rated as Low vs. High by each model (split by the model's median difficulty).}
    \label{tab:difficulty_control_means}
    \centering
    \begin{tabular}{lccc|ccc}
        \toprule
        \multirow{2}{*}{\textbf{\texttt{\textit{ModelName}-Dif}}} & \multicolumn{3}{c|}{\texttt{Experts-Dif} (1–5)} & \multicolumn{3}{c}{\texttt{Students-Dif} (0–1)} \\
        \cmidrule(lr){2-4} \cmidrule(lr){5-7}
        & Low & High & \textit{p} & Low & High & \textit{p} \\
        \midrule
        \texttt{GPT-4o-Dif} & 2.03& 2.96& .00006& .13& .31& .0004\\
        \texttt{Gemma-2-Dif} & 2.18& 2.42& .2502& .19& .26& .2475\\
        \texttt{Ptt5-v2+Gemma-2-Dif} & 2.12& 2.85& .0073& .13& .28& .0404\\
        \bottomrule
    \end{tabular}
\end{table}

The results show a consistent trend: mean difficulty values are lower for the Low group and higher for the High group.
This pattern holds across both \texttt{Experts-Dif} and \texttt{Students-Dif} scores, with statistically significant differences observed for \texttt{GPT-4o-Dif} and \texttt{Ptt5-v2+Gemma-2-Dif}.
These findings suggest that the difficulty values assigned by the models are meaningfully associated with the actual difficulty of the generated MCQs --- that is, when prompted to generate questions with varying difficulty, the resulting MCQs exhibit differences that align with both expert and student perceptions.

\subsection{Feature Influence on Difficulty Perception} \label{sec:difficulty_out_perspective}

This section analyzes how specific features of MCQs influence difficulty as perceived by experts, students, and models. Specifically, we aim to understand whether certain MCQ characteristics systematically affect how difficulty is perceived.

\subsubsection{Features}
We examine surface and semantic features across the difficulty perspectives (\texttt{Experts-Dif}, \texttt{Students-Dif}, and \texttt{\textit{ModelName}-Dif}):

\begin{itemize}
    \item \texttt{Question-Size}: The number of tokens in the \textit{wh}-question.
    \item \texttt{Text-Size}: The number of tokens in the narrative text.
    \item \texttt{Sem-Sim-Correct-Distr}: The average semantic similarity\footnote{Semantic similarity is computed using the Portuguese \texttt{Serafim} Sentence Encoder Model~\citep{rodrigues_2023_albertina}.} between the MCQ correct answer and each distractor. Higher values indicate that the correct answer is semantically closer to the distractors.
    \item \texttt{Sem-Sim-Options}: The average semantic similarity between each pair of MCQ options. Higher values indicate greater semantic similarity between options.
    \item \texttt{Sem-Sim-Question-Options}: The average semantic similarity between the \textit{wh}-question and each option. Higher values indicate that the question is semantically closer to the options.
\end{itemize}

To analyze these features, we divide the MCQs into two groups for each feature independently, based on the distribution of its values:
\begin{itemize}
\item \textbf{Low Group}: MCQs with values below the median for that specific feature.
\item \textbf{High Group}: MCQs with values above the median for that specific feature.
\end{itemize}

\subsubsection{Results}

Table~\ref{tab:metrics_corr} presents the mean difficulty values from all perspectives for each feature, comparing MCQs in the Low and High groups. We observed the following:

\noindent \textbf{Question-Size:} Shorter questions (Low Group) are generally perceived as easier across most perspectives, with a statistically significant difference observed in \texttt{Experts-Dif} ($p < 0.05$).\\
\noindent \textbf{Text-Size:} This feature shows no consistent effect on perceived difficulty across perspectives.\\
\noindent \textbf{Semantic similarity features:} Higher semantic similarity (High groups) tends to increase difficulty perception. This effect is particularly clear for the \texttt{Sem-Sim-Question-Options} feature in both \texttt{Students-Dif} and \texttt{GPT-4o-Dif} ($p < 0.05$), suggesting that MCQs are harder when the \textit{wh}-question is semantically similar to the answer options.

\begin{sidewaystable}
    \caption{Mean difficulty values from each perspective (\texttt{-Dif}) for MCQs grouped into \textbf{Low} and \textbf{High} groups (split by the median value of each feature).}
    \label{tab:metrics_corr}
    \centering
    \setlength{\tabcolsep}{2pt} 
    \begin{tabular*}{\textheight}{@{\hskip 2pt}l ccc@{\hskip 2pt} ccc@{\hskip 2pt} ccc@{\hskip 2pt} ccc@{\hskip 2pt} ccc@{\hskip 2pt}}
        \toprule
        Feature & \multicolumn{3}{c}{\texttt{Experts-Dif} (1-5)} & \multicolumn{3}{c}{\texttt{Students-Dif} (0-1)} & \multicolumn{3}{c}{ \texttt{GPT-4o-Dif} (0-100)} & \multicolumn{3}{c}{\texttt{Gemma-2-Dif} (0-100)} & \multicolumn{3}{c}{\texttt{Ptt5-v2+Gemma-2-Dif} (0-100)} \\
        \cmidrule(lr){2-4} \cmidrule(lr){5-7} \cmidrule(lr){8-10} \cmidrule(lr){11-13} \cmidrule(lr){14-16}
        & Low & High & p & Low & High & p & Low & High & p & Low & High & p & Low & High & p \\
        \midrule
        \texttt{Question-Size} & 2.48 & 2.37 & .274 & .20 & .23 & .571 & 27.22 & 30.21 & .612 & 31.88 & 35.88 & .636 & 34.09 & 38.21 & .614 \\
        \texttt{Text-Size} & 2.63 & 2.26 & .001 & .217 & .218 & .898 & 30.88 & 27.81 & .414 & 36.56 & 31.47 & .456 & 41.67 & 31.54 & .081 \\
        \texttt{Sem-Sim-Correct-Distr} & 2.35 & 2.49 & .269 & .195 & .240 & .247 & 26.56 & 32.06 & .291 & 29.38 & 38.24 & .179 & 37.08 & 35.77 & .872 \\
        \texttt{Sem-Sim-Options} & 2.32 & 2.52 & .150 & .190 & .245 & .181 & 25.63 & 32.94 & .157 & 32.81 & 35.00 & .490 & 35.42 & 37.31 & .978 \\
        \texttt{Sem-Sim-Question-Options} & 2.30 & 2.54 & .071 & .182 & .253 & .029 & 22.50 & 35.88 & .007 & 30.00 & 37.65 & .106 & 33.75 & 38.85 & .530 \\
        \bottomrule
    \end{tabular*}
\end{sidewaystable}

Additionally, Table~\ref{tab:narrative_corr} presents mean difficulty values for each narrative element. MCQs targeting \texttt{character} are generally perceived as easier, while those focused on \texttt{causal relationship} are perceived as more difficult. This pattern is particularly pronounced in the \texttt{Experts-Dif} and \texttt{Ptt5-v2-Gemma-2-Dif} perspectives, where the differences are statistically significant ($p < 0.05$).
This can be explained by the nature of these questions: \texttt{causal relationship} questions often begin with ``Why'' and correspond to higher-order thinking skills in Bloom's taxonomy~\citep{bloom_2002_revised}, whereas \texttt{character}-based questions tend to require lower-level recall or identification.

\begin{table}[!ht]
    \caption{Mean difficulty values (\texttt{-Dif}) for each narrative element across different difficulty perspectives.}
    \label{tab:narrative_corr}
    \centering
    \begin{tabular*}{\textwidth}{@{\extracolsep\fill}l ccc ccc}
        \toprule
        \textbf{Narrative Element} & \texttt{Experts-Dif} & \texttt{Students-Dif} & \texttt{GPT-4o-Dif} & \texttt{Gemma-2-Dif} & \texttt{P.+G.-Dif}\footnotemark[1] \\
        \midrule
        \texttt{Character} & 2.07 & .170 & 23.33 & 27.00 & 18.57 \\
        \texttt{Setting} & 2.41 & .218 & 32.69 & 29.55 & 28.57 \\
        \texttt{Action} & 2.37 & .203 & 16.50 & 28.75 & 40.00 \\
        \texttt{Feeling} & 2.53 & .258 & 22.50 & 38.13 & 45.00 \\
        \texttt{Causal Relationships} & 2.71 & .242 & 38.58 & 48.00 & 60.00 \\
        \midrule
        \textbf{p-value} & .\textbf{014} & .527 & .134 & .183 & \textbf{.002} \\
        \bottomrule
    \end{tabular*}
    \footnotetext[1]{\texttt{Ptt5-v2+Gemma-2-Dif}}
\end{table}

\subsection{Discussion of Results}

We revisit our third research question (RQ3): \textit{How do the perceived difficulty characteristics of generated MCQs compare across expert review, psychometric properties, and model-annotated difficulty values?}

We found that models predicting difficulty \textit{after} generating the full MCQ tend to align better with human-perceived difficulty. This highlights a key insight: generating MCQs while assigning difficulty simultaneously (within a single prompt) may reduce the accuracy of difficulty estimation.
We also observed a weak but noticeable correlation between experts and students, suggesting that their perceptions are only partially aligned. This is reflected in the narrative element analysis, where certain categories significantly influence expert perceptions of difficulty but have a limited impact on student perceptions.

Interestingly, model-annotated difficulty values correlate more strongly with experts' perception than with students'. This implies that LLMs may be more aligned with experts' difficulty perception than with actual student performance.

Despite most MCQs being perceived as relatively easy by both experts and students, our results show that models, particularly \texttt{GPT-4o} and \texttt{Ptt5-v2+Gemma-2}, can effectively assign difficulty values that correspond to actual differences in perceived difficulty.

Finally, we find some shared indicators of difficulty across all perspectives, such as question size and semantic similarity within MCQ components.
For example, higher semantic similarity between the \textit{wh}-question and its options tends to make MCQs perceived as more difficult.
This may be because when all options appear closely related to the \textit{wh}-question, it becomes harder to discriminate the correct answer.
Surprisingly, the semantic similarity between correct answers and distractors, a popular feature in difficulty estimation~\citep{kurdi_2020_education}, showed no consistent relationship with perceived difficulty.

\section{Summary of Findings} \label{sec:summary_findings}

This study evaluated the quality of reading comprehension MCQs generated for Portuguese elementary students using expert review, psychometric properties, and perceived difficulty analysis.
Our findings offer insights that can inform broader educational MCQ generation, particularly for those applying LLMs in lower-resourced language contexts. Key takeaways include:
\begin{itemize}
    \item \textit{Wh}-questions generated by LLMs are generally comparable to human-authored ones regarding well-formedness.
    \item While well-formed, some generated \textit{wh}-questions appear in unintended variants (e.g., Brazilian Portuguese), highlighting both prompt adherence issues and the prevalence of specific variants in LLM training data.
    \item Although rare, semantic issues can occur, with LLMs occasionally producing unclear or nonsensical questions. However, these problems occur similarly in human-authored MCQs.
    \item Generated MCQs typically include clear options, though some require rewording for improved clarity. Again, these issues are also found in human-authored MCQs.
    \item The general difficulty and plausibility of options do not differ significantly between generated and human-authored MCQs.
    \item Psychometric properties (difficulty and discrimination) of generated MCQs are on par with human-authored ones. However, students tend to engage more with distractors in human-authored MCQs, suggesting they have higher plausibility. Additionally, human-authored MCQs better adhere to option selection principles (3-rule test).
    \item A two-step generation method (first generating a \textit{wh}-question, then the full MCQ) generally yields lower-quality questions and results in lower difficulty and discrimination scores.
    \item LLMs can effectively generate MCQs that vary in difficulty levels and narrative elements.
    \item Model-annotated difficulty correlates more strongly with human perception (experts and students) when it is estimated after the full MCQ is generated.
    \item Perceived MCQs difficulty from LLMs aligns more closely with expert perceptions than with those of students.
    \item Features such as narrative category, question length, and semantic similarity are shared indicators across experts, students, and models in shaping perceived MCQ difficulty.
\end{itemize}

\section{Limitations} \label{sec:limitations}

This study presents some limitations that should be considered when interpreting the results:

\begin{itemize}
    \item \textbf{Limited Language Generalizability}: This study focuses specifically on generating reading comprehension MCQs for Portuguese elementary students. This may limit the generalizability of the findings to other languages or Portuguese variants. Still, we believe the study is particularly relevant for those applying LLMs in underrepresented or lower-resourced language settings.
    \item \textbf{Limited Exploration of Prompt Design}: This study adopts a fixed prompt structure rather than systematically comparing prompting strategies. Exploring different prompt designs (as explored by \citet{scaria_2024_aied}) could offer deeper insights into improving the quality of generated MCQs.
    \item \textbf{Limited Comparison of LLMs}: We evaluate two specific LLMs in the one-step generation method, selected from an initial comparison. While many other models (proprietary and open-source) could be explored, expanding the comparison would require evaluating more MCQs per model. Given the effort required from both experts and students, we intentionally limited the number of models to ensure a manageable and in-depth evaluation.
    \item \textbf{Restricted Understanding of Two-Step Method Performance}: The two-step generation method was implemented using model \texttt{Ptt5-v2} for step one and \texttt{Gemma-2} for step two. This approach yielded lower performance than the one-step method with LLMs. However, this may be due to the first-step model rather than the approach itself. \texttt{Ptt5-v2} is a fine-tuned smaller ($<$1B parameters) model, which is open-source and fast, but may act as a bottleneck in the overall quality of the generated MCQs.
\end{itemize}

\section{Conclusions}\label{sec:conclusions}

This study evaluated the quality of reading comprehension MCQs generated by LLMs for Portuguese elementary students.
We found that current LLMs can produce well-formed MCQs comparable to human-authored ones. Common issues include occasional generation of \textit{wh}-questions in unintended language variants, semantic flaws, and ambiguous answerability. Notably, the latter two issues are also observed in human-authored MCQs, highlighting the similarity in terms of quality.
LLMs demonstrated the ability to control narrative attributes embedded in MCQs and are effective in assigning difficulty levels that align with human perceived difficulty --- particularly when difficulty is predicted after the full MCQ is generated.
However, psychometric analysis revealed that distractors in human-authored MCQs engage students more and better adhere to best practices for option selection.
Overall, we consider this study a reference point for the use of current LLMs in generating MCQs for reading comprehension, especially in less-resourced languages like Portuguese. It highlights both the potential and the limitations of these models, providing a valuable benchmark for future research. We hope this work will serve as a useful comparison for others investigating automatic question generation and its applications in education.

For future work, we aim to investigate the impact of generated MCQs on learning outcomes and knowledge retention. While expert reviews and psychometric analysis offer valuable insights on one-time assessments, long-term classroom deployment would provide a deeper understanding of their educational effectiveness.
This brings challenges such as variability in teaching contexts and learner engagement around automated content.
Additionally, exploring personalized QG, by tailoring difficulty to individual learners, presents a promising avenue for more adaptive and effective assessment.






\backmatter





\bmhead{Acknowledgements}

The authors would like to thank \textit{Porto Editora} for supporting this research.
We thank the collaborators from the \textit{Divis\~ao de Tecnologias Educativas da Porto Editora} who participated in the expert evaluation process.  
Our gratitude extends to the teachers who voluntarily took part in the expert evaluation.
We would like to specifically acknowledge those who agreed to share their names: Adalgisa Nunes, Carina Castro, Cecília Carvalho, Dalila Carvalhais, Joaquina Reis, Luísa Loureiro, Neuza Pinto and Sandra Monteiro.
We also thank the administration of \textit{Agrupamento de Escolas de S~ao Louren\c{c}o} for authorizing the evaluation process with students, as well as all the teachers who facilitated the evaluation process with their students in class.
Finally, we thank \textit{Dire\c{c}\~ao-Geral da Educa\c{c}\~ao} for authorizing the evaluation process with students at schools.
The student inquiry authorization has reference 1305300002\footnote{https://mime.dgeec.mec.pt/InqueritoConsultar.aspx?id=16148}.


\section*{Declarations}

\subsection*{Funding}

This work was financially supported by UID/00027 -- the Artificial Intelligence and
Computer Science Laboratory (LIACC), funded by \textit{Funda\c{c}\~{a}o para a Ci\^{e}ncia e a Tecnologia} (FCT), I.P./ MCTES through national funds.
Bernardo Leite is supported by a PhD studentship (with reference 2021.05432.BD), funded by FCT.

\bibliography{sn-bibliography}


\begin{thebibliography}{66}
\ifx \bisbn   \undefined \def \bisbn  #1{ISBN #1}\fi
\ifx \binits  \undefined \def \binits#1{#1}\fi
\ifx \bauthor  \undefined \def \bauthor#1{#1}\fi
\ifx \batitle  \undefined \def \batitle#1{#1}\fi
\ifx \bjtitle  \undefined \def \bjtitle#1{#1}\fi
\ifx \bvolume  \undefined \def \bvolume#1{\textbf{#1}}\fi
\ifx \byear  \undefined \def \byear#1{#1}\fi
\ifx \bissue  \undefined \def \bissue#1{#1}\fi
\ifx \bfpage  \undefined \def \bfpage#1{#1}\fi
\ifx \blpage  \undefined \def \blpage #1{#1}\fi
\ifx \burl  \undefined \def \burl#1{\textsf{#1}}\fi
\ifx \doiurl  \undefined \def \doiurl#1{\url{https://doi.org/#1}}\fi
\ifx \betal  \undefined \def \betal{\textit{et al.}}\fi
\ifx \binstitute  \undefined \def \binstitute#1{#1}\fi
\ifx \binstitutionaled  \undefined \def \binstitutionaled#1{#1}\fi
\ifx \bctitle  \undefined \def \bctitle#1{#1}\fi
\ifx \beditor  \undefined \def \beditor#1{#1}\fi
\ifx \bpublisher  \undefined \def \bpublisher#1{#1}\fi
\ifx \bbtitle  \undefined \def \bbtitle#1{#1}\fi
\ifx \bedition  \undefined \def \bedition#1{#1}\fi
\ifx \bseriesno  \undefined \def \bseriesno#1{#1}\fi
\ifx \blocation  \undefined \def \blocation#1{#1}\fi
\ifx \bsertitle  \undefined \def \bsertitle#1{#1}\fi
\ifx \bsnm \undefined \def \bsnm#1{#1}\fi
\ifx \bsuffix \undefined \def \bsuffix#1{#1}\fi
\ifx \bparticle \undefined \def \bparticle#1{#1}\fi
\ifx \barticle \undefined \def \barticle#1{#1}\fi
\bibcommenthead
\ifx \bconfdate \undefined \def \bconfdate #1{#1}\fi
\ifx \botherref \undefined \def \botherref #1{#1}\fi
\ifx \url \undefined \def \url#1{\textsf{#1}}\fi
\ifx \bchapter \undefined \def \bchapter#1{#1}\fi
\ifx \bbook \undefined \def \bbook#1{#1}\fi
\ifx \bcomment \undefined \def \bcomment#1{#1}\fi
\ifx \oauthor \undefined \def \oauthor#1{#1}\fi
\ifx \citeauthoryear \undefined \def \citeauthoryear#1{#1}\fi
\ifx \endbibitem  \undefined \def \endbibitem {}\fi
\ifx \bconflocation  \undefined \def \bconflocation#1{#1}\fi
\ifx \arxivurl  \undefined \def \arxivurl#1{\textsf{#1}}\fi
\csname PreBibitemsHook\endcsname

\bibitem[\protect\citeauthoryear{Das et~al.}{2021}]{das_2021_objective}
\begin{barticle}
\bauthor{\bsnm{Das}, \binits{B.}},
\bauthor{\bsnm{Majumder}, \binits{M.}},
\bauthor{\bsnm{Phadikar}, \binits{S.}},
\bauthor{\bsnm{Sekh}, \binits{A.A.}}:
\batitle{Automatic question generation and answer assessment: a survey}.
\bjtitle{Research and Practice in Technology Enhanced Learning}
\bvolume{16}(\bissue{1}),
\bfpage{5}
(\byear{2021})
\end{barticle}
\endbibitem

\bibitem[\protect\citeauthoryear{Kurdi et~al.}{2020}]{kurdi_2020_education}
\begin{barticle}
\bauthor{\bsnm{Kurdi}, \binits{G.}},
\bauthor{\bsnm{Leo}, \binits{J.}},
\bauthor{\bsnm{Parsia}, \binits{B.}},
\bauthor{\bsnm{Sattler}, \binits{U.}},
\bauthor{\bsnm{Al-Emari}, \binits{S.}}:
\batitle{A systematic review of automatic question generation for educational purposes}.
\bjtitle{International Journal of Artificial Intelligence in Education}
\bvolume{30}(\bissue{1}),
\bfpage{121}--\blpage{204}
(\byear{2020})
\end{barticle}
\endbibitem

\bibitem[\protect\citeauthoryear{CH and Saha}{2020}]{ch_2020_mcq_survey}
\begin{barticle}
\bauthor{\bsnm{CH}, \binits{D.R.}},
\bauthor{\bsnm{Saha}, \binits{S.K.}}:
\batitle{Automatic multiple choice question generation from text: A survey}.
\bjtitle{IEEE Transactions on Learning Technologies}
\bvolume{13}(\bissue{1}),
\bfpage{14}--\blpage{25}
(\byear{2020})
\doiurl{10.1109/TLT.2018.2889100}
\end{barticle}
\endbibitem

\bibitem[\protect\citeauthoryear{Alhazmi et~al.}{2024}]{alhazmi_2024_distractor_survey}
\begin{bchapter}
\bauthor{\bsnm{Alhazmi}, \binits{E.}},
\bauthor{\bsnm{Sheng}, \binits{Q.Z.}},
\bauthor{\bsnm{Zhang}, \binits{W.E.}},
\bauthor{\bsnm{Zaib}, \binits{M.}},
\bauthor{\bsnm{Alhazmi}, \binits{A.}}:
\bctitle{Distractor generation in multiple-choice tasks: A survey of methods, datasets, and evaluation}.
In: \beditor{\bsnm{Al-Onaizan}, \binits{Y.}},
\beditor{\bsnm{Bansal}, \binits{M.}},
\beditor{\bsnm{Chen}, \binits{Y.-N.}} (eds.)
\bbtitle{Proceedings of the 2024 Conference on Empirical Methods in Natural Language Processing},
pp. \bfpage{14437}--\blpage{14458}.
\bpublisher{Association for Computational Linguistics},
\blocation{Miami, Florida, USA}
(\byear{2024}).
\doiurl{10.18653/v1/2024.emnlp-main.799} .
\burl{https://aclanthology.org/2024.emnlp-main.799/}
\end{bchapter}
\endbibitem

\bibitem[\protect\citeauthoryear{Lee et~al.}{2024}]{lee_2024_few_shot_is_enough}
\begin{barticle}
\bauthor{\bsnm{Lee}, \binits{U.}},
\bauthor{\bsnm{Jung}, \binits{H.}},
\bauthor{\bsnm{Jeon}, \binits{Y.}},
\bauthor{\bsnm{Sohn}, \binits{Y.}},
\bauthor{\bsnm{Hwang}, \binits{W.}},
\bauthor{\bsnm{Moon}, \binits{J.}},
\bauthor{\bsnm{Kim}, \binits{H.}}:
\batitle{Few-shot is enough: exploring chatgpt prompt engineering method for automatic question generation in english education}.
\bjtitle{Education and Information Technologies}
\bvolume{29}(\bissue{9}),
\bfpage{11483}--\blpage{11515}
(\byear{2024})
\end{barticle}
\endbibitem

\bibitem[\protect\citeauthoryear{Wang et~al.}{2022}]{wang_2022_towards_modular}
\begin{bchapter}
\bauthor{\bsnm{Wang}, \binits{X.}},
\bauthor{\bsnm{Fan}, \binits{S.}},
\bauthor{\bsnm{Houghton}, \binits{J.}},
\bauthor{\bsnm{Wang}, \binits{L.}}:
\bctitle{Towards process-oriented, modular, and versatile question generation that meets educational needs}.
In: \beditor{\bsnm{Carpuat}, \binits{M.}},
\beditor{\bsnm{Marneffe}, \binits{M.-C.}},
\beditor{\bsnm{Meza~Ruiz}, \binits{I.V.}} (eds.)
\bbtitle{Proceedings of the 2022 Conference of the North American Chapter of the Association for Computational Linguistics: Human Language Technologies},
pp. \bfpage{291}--\blpage{302}.
\bpublisher{Association for Computational Linguistics},
\blocation{Seattle, United States}
(\byear{2022}).
\doiurl{10.18653/v1/2022.naacl-main.22} .
\burl{https://aclanthology.org/2022.naacl-main.22/}
\end{bchapter}
\endbibitem

\bibitem[\protect\citeauthoryear{Holstein and Aleven}{2022}]{holstein_aleven_2022_fallible_ai}
\begin{barticle}
\bauthor{\bsnm{Holstein}, \binits{K.}},
\bauthor{\bsnm{Aleven}, \binits{V.}}:
\batitle{Designing for human–ai complementarity in k-12 education}.
\bjtitle{AI Magazine}
\bvolume{43}(\bissue{2}),
\bfpage{239}--\blpage{248}
(\byear{2022})
\doiurl{10.1002/aaai.12058}
\end{barticle}
\endbibitem

\bibitem[\protect\citeauthoryear{Alsubait et~al.}{2016}]{alsubait_2016_ontology}
\begin{barticle}
\bauthor{\bsnm{Alsubait}, \binits{T.}},
\bauthor{\bsnm{Parsia}, \binits{B.}},
\bauthor{\bsnm{Sattler}, \binits{U.}}:
\batitle{Ontology-based multiple choice question generation}.
\bjtitle{KI-K{\"u}nstliche Intelligenz}
\bvolume{30},
\bfpage{183}--\blpage{188}
(\byear{2016})
\end{barticle}
\endbibitem

\bibitem[\protect\citeauthoryear{K{\i}yak et~al.}{2024}]{kiyak_2024_chatgpt_medical}
\begin{barticle}
\bauthor{\bsnm{K{\i}yak}, \binits{Y.S.}},
\bauthor{\bsnm{Co{\c{s}}kun}, \binits{{\"O}.}},
\bauthor{\bsnm{Budako{\u{g}}lu}, \binits{I.{\. I}.}},
\bauthor{\bsnm{Uluo{\u{g}}lu}, \binits{C.}}:
\batitle{Chatgpt for generating multiple-choice questions: evidence on the use of artificial intelligence in automatic item generation for a rational pharmacotherapy exam}.
\bjtitle{European journal of clinical pharmacology}
\bvolume{80}(\bissue{5}),
\bfpage{729}--\blpage{735}
(\byear{2024})
\end{barticle}
\endbibitem

\bibitem[\protect\citeauthoryear{Malec}{2024}]{malec_2024_csedu_mcq_vocabulary}
\begin{bchapter}
\bauthor{\bsnm{Malec}, \binits{W.}}:
\bctitle{Investigating the Quality of AI-Generated Distractors for a Multiple-Choice Vocabulary Test}.
In: \bbtitle{Proceedings of the 16th International Conference on Computer Supported Education - Volume 1: AIG},
pp. \bfpage{836}--\blpage{843}.
\bpublisher{SciTePress}, \blocation{???}
(\byear{2024}).
\doiurl{10.5220/0012762400003693} .
\bcomment{INSTICC}
\end{bchapter}
\endbibitem

\bibitem[\protect\citeauthoryear{Ashok~Kumar et~al.}{2023}]{ashok_2023_rc}
\begin{bchapter}
\bauthor{\bsnm{Ashok~Kumar}, \binits{N.}},
\bauthor{\bsnm{Fernandez}, \binits{N.}},
\bauthor{\bsnm{Wang}, \binits{Z.}},
\bauthor{\bsnm{Lan}, \binits{A.}}:
\bctitle{Improving reading comprehension question generation with data augmentation and overgenerate-and-rank}.
In: \beditor{\bsnm{Kochmar}, \binits{E.}},
\beditor{\bsnm{Burstein}, \binits{J.}},
\beditor{\bsnm{Horbach}, \binits{A.}},
\beditor{\bsnm{Laarmann-Quante}, \binits{R.}},
\beditor{\bsnm{Madnani}, \binits{N.}},
\beditor{\bsnm{Tack}, \binits{A.}},
\beditor{\bsnm{Yaneva}, \binits{V.}},
\beditor{\bsnm{Yuan}, \binits{Z.}},
\beditor{\bsnm{Zesch}, \binits{T.}} (eds.)
\bbtitle{Proceedings of the 18th Workshop on Innovative Use of NLP for Building Educational Applications (BEA 2023)},
pp. \bfpage{247}--\blpage{259}.
\bpublisher{Association for Computational Linguistics},
\blocation{Toronto, Canada}
(\byear{2023}).
\doiurl{10.18653/v1/2023.bea-1.22} .
\burl{https://aclanthology.org/2023.bea-1.22/}
\end{bchapter}
\endbibitem

\bibitem[\protect\citeauthoryear{Sim and Berthelsen}{2014}]{sim_2014_rc}
\begin{barticle}
\bauthor{\bsnm{Sim}, \binits{S.}},
\bauthor{\bsnm{Berthelsen}, \binits{D.}}:
\batitle{Shared book reading by parents with young children: Evidence-based practice}.
\bjtitle{Australasian Journal of Early Childhood}
\bvolume{39}(\bissue{1}),
\bfpage{50}--\blpage{55}
(\byear{2014})
\end{barticle}
\endbibitem

\bibitem[\protect\citeauthoryear{Lynch et~al.}{2008}]{lynch_2008_rc}
\begin{barticle}
\bauthor{\bsnm{Lynch}, \binits{J.S.}},
\bauthor{\bsnm{Van Den~Broek}, \binits{P.}},
\bauthor{\bsnm{Kremer}, \binits{K.E.}},
\bauthor{\bsnm{Kendeou}, \binits{P.}},
\bauthor{\bsnm{White}, \binits{M.J.}},
\bauthor{\bsnm{Lorch}, \binits{E.P.}}:
\batitle{The development of narrative comprehension and its relation to other early reading skills}.
\bjtitle{Reading Psychology}
\bvolume{29}(\bissue{4}),
\bfpage{327}--\blpage{365}
(\byear{2008})
\end{barticle}
\endbibitem

\bibitem[\protect\citeauthoryear{Xu et~al.}{2022}]{xu_2022_fairytaleqa}
\begin{bchapter}
\bauthor{\bsnm{Xu}, \binits{Y.}},
\bauthor{\bsnm{Wang}, \binits{D.}},
\bauthor{\bsnm{Yu}, \binits{M.}},
\bauthor{\bsnm{Ritchie}, \binits{D.}},
\bauthor{\bsnm{Yao}, \binits{B.}},
\bauthor{\bsnm{Wu}, \binits{T.}},
\bauthor{\bsnm{Zhang}, \binits{Z.}},
\bauthor{\bsnm{Li}, \binits{T.}},
\bauthor{\bsnm{Bradford}, \binits{N.}},
\bauthor{\bsnm{Sun}, \binits{B.}},
\bauthor{\bsnm{Hoang}, \binits{T.}},
\bauthor{\bsnm{Sang}, \binits{Y.}},
\bauthor{\bsnm{Hou}, \binits{Y.}},
\bauthor{\bsnm{Ma}, \binits{X.}},
\bauthor{\bsnm{Yang}, \binits{D.}},
\bauthor{\bsnm{Peng}, \binits{N.}},
\bauthor{\bsnm{Yu}, \binits{Z.}},
\bauthor{\bsnm{Warschauer}, \binits{M.}}:
\bctitle{Fantastic questions and where to find them: {F}airytale{QA} {--} an authentic dataset for narrative comprehension}.
In: \bbtitle{Proceedings of the 60th Annual Meeting of the Association for Computational Linguistics (Volume 1: Long Papers)},
pp. \bfpage{447}--\blpage{460}.
\bpublisher{Association for Computational Linguistics},
\blocation{Dublin, Ireland}
(\byear{2022}).
\doiurl{10.18653/v1/2022.acl-long.34}
\end{bchapter}
\endbibitem

\bibitem[\protect\citeauthoryear{Eo et~al.}{2023}]{eo_2023_diversity}
\begin{bchapter}
\bauthor{\bsnm{Eo}, \binits{S.}},
\bauthor{\bsnm{Moon}, \binits{H.}},
\bauthor{\bsnm{Kim}, \binits{J.}},
\bauthor{\bsnm{Hur}, \binits{Y.}},
\bauthor{\bsnm{Kim}, \binits{J.}},
\bauthor{\bsnm{Lee}, \binits{S.}},
\bauthor{\bsnm{Chun}, \binits{C.}},
\bauthor{\bsnm{Park}, \binits{S.}},
\bauthor{\bsnm{Lim}, \binits{H.}}:
\bctitle{Towards diverse and effective question-answer pair generation from children storybooks}.
In: \beditor{\bsnm{Rogers}, \binits{A.}},
\beditor{\bsnm{Boyd-Graber}, \binits{J.}},
\beditor{\bsnm{Okazaki}, \binits{N.}} (eds.)
\bbtitle{Findings of the Association for Computational Linguistics: ACL 2023},
pp. \bfpage{6100}--\blpage{6115}.
\bpublisher{Association for Computational Linguistics},
\blocation{Toronto, Canada}
(\byear{2023}).
\doiurl{10.18653/v1/2023.findings-acl.380} .
\burl{https://aclanthology.org/2023.findings-acl.380/}
\end{bchapter}
\endbibitem

\bibitem[\protect\citeauthoryear{Alhazmi et~al.}{2024}]{alhazmi_2024_survey_mcq}
\begin{bchapter}
\bauthor{\bsnm{Alhazmi}, \binits{E.}},
\bauthor{\bsnm{Sheng}, \binits{Q.Z.}},
\bauthor{\bsnm{Zhang}, \binits{W.E.}},
\bauthor{\bsnm{Zaib}, \binits{M.}},
\bauthor{\bsnm{Alhazmi}, \binits{A.}}:
\bctitle{Distractor generation in multiple-choice tasks: A survey of methods, datasets, and evaluation}.
In: \beditor{\bsnm{Al-Onaizan}, \binits{Y.}},
\beditor{\bsnm{Bansal}, \binits{M.}},
\beditor{\bsnm{Chen}, \binits{Y.-N.}} (eds.)
\bbtitle{Proceedings of the 2024 Conference on Empirical Methods in Natural Language Processing},
pp. \bfpage{14437}--\blpage{14458}.
\bpublisher{Association for Computational Linguistics},
\blocation{Miami, Florida, USA}
(\byear{2024}).
\doiurl{10.18653/v1/2024.emnlp-main.799} .
\burl{https://aclanthology.org/2024.emnlp-main.799/}
\end{bchapter}
\endbibitem

\bibitem[\protect\citeauthoryear{Jiang and Lee}{2017}]{jiang_2017_dist}
\begin{bchapter}
\bauthor{\bsnm{Jiang}, \binits{S.}},
\bauthor{\bsnm{Lee}, \binits{J.}}:
\bctitle{Distractor generation for {C}hinese fill-in-the-blank items}.
In: \beditor{\bsnm{Tetreault}, \binits{J.}},
\beditor{\bsnm{Burstein}, \binits{J.}},
\beditor{\bsnm{Leacock}, \binits{C.}},
\beditor{\bsnm{Yannakoudakis}, \binits{H.}} (eds.)
\bbtitle{Proceedings of the 12th Workshop on Innovative Use of {NLP} for Building Educational Applications},
pp. \bfpage{143}--\blpage{148}.
\bpublisher{Association for Computational Linguistics},
\blocation{Copenhagen, Denmark}
(\byear{2017}).
\doiurl{10.18653/v1/W17-5015} .
\burl{https://aclanthology.org/W17-5015/}
\end{bchapter}
\endbibitem

\bibitem[\protect\citeauthoryear{Susanti et~al.}{2015}]{susanti_2015_dist}
\begin{bchapter}
\bauthor{\bsnm{Susanti}, \binits{Y.}},
\bauthor{\bsnm{Iida}, \binits{R.}},
\bauthor{\bsnm{Tokunaga}, \binits{T.}}:
\bctitle{Automatic Generation of English Vocabulary Tests}.
In: \bbtitle{Proceedings of the 7th International Conference on Computer Supported Education - CSEDU},
pp. \bfpage{77}--\blpage{87}.
\bpublisher{SciTePress}, \blocation{???}
(\byear{2015}).
\doiurl{10.5220/0005437200770087} .
\bcomment{INSTICC}
\end{bchapter}
\endbibitem

\bibitem[\protect\citeauthoryear{Pino and Eskenazi}{2009}]{pino_2009_dist}
\begin{bchapter}
\bauthor{\bsnm{Pino}, \binits{J.}},
\bauthor{\bsnm{Eskenazi}, \binits{M.}}:
\bctitle{Semi-automatic generation of cloze question distractors effect of students' l1.}
In: \bbtitle{SLaTE},
pp. \bfpage{65}--\blpage{68}
(\byear{2009})
\end{bchapter}
\endbibitem

\bibitem[\protect\citeauthoryear{Lai et~al.}{2017}]{lai_2017_race}
\begin{bchapter}
\bauthor{\bsnm{Lai}, \binits{G.}},
\bauthor{\bsnm{Xie}, \binits{Q.}},
\bauthor{\bsnm{Liu}, \binits{H.}},
\bauthor{\bsnm{Yang}, \binits{Y.}},
\bauthor{\bsnm{Hovy}, \binits{E.}}:
\bctitle{{RACE}: Large-scale {R}e{A}ding comprehension dataset from examinations}.
In: \beditor{\bsnm{Palmer}, \binits{M.}},
\beditor{\bsnm{Hwa}, \binits{R.}},
\beditor{\bsnm{Riedel}, \binits{S.}} (eds.)
\bbtitle{Proceedings of the 2017 Conference on Empirical Methods in Natural Language Processing},
pp. \bfpage{785}--\blpage{794}.
\bpublisher{Association for Computational Linguistics},
\blocation{Copenhagen, Denmark}
(\byear{2017}).
\doiurl{10.18653/v1/D17-1082} .
\burl{https://aclanthology.org/D17-1082/}
\end{bchapter}
\endbibitem

\bibitem[\protect\citeauthoryear{Gao et~al.}{2019}]{gao_2019_race}
\begin{barticle}
\bauthor{\bsnm{Gao}, \binits{Y.}},
\bauthor{\bsnm{Bing}, \binits{L.}},
\bauthor{\bsnm{Li}, \binits{P.}},
\bauthor{\bsnm{King}, \binits{I.}},
\bauthor{\bsnm{Lyu}, \binits{M.R.}}:
\batitle{Generating distractors for reading comprehension questions from real examinations}.
\bjtitle{Proceedings of the AAAI Conference on Artificial Intelligence}
\bvolume{33}(\bissue{01}),
\bfpage{6423}--\blpage{6430}
(\byear{2019})
\doiurl{10.1609/aaai.v33i01.33016423}
\end{barticle}
\endbibitem

\bibitem[\protect\citeauthoryear{Zhou et~al.}{2020}]{zhou_2020_race}
\begin{barticle}
\bauthor{\bsnm{Zhou}, \binits{X.}},
\bauthor{\bsnm{Luo}, \binits{S.}},
\bauthor{\bsnm{Wu}, \binits{Y.}}:
\batitle{Co-attention hierarchical network: Generating coherent long distractors for reading comprehension}.
\bjtitle{Proceedings of the AAAI Conference on Artificial Intelligence}
\bvolume{34}(\bissue{05}),
\bfpage{9725}--\blpage{9732}
(\byear{2020})
\doiurl{10.1609/aaai.v34i05.6522}
\end{barticle}
\endbibitem

\bibitem[\protect\citeauthoryear{Raffel et~al.}{2020}]{raffel_2020_t5}
\begin{barticle}
\bauthor{\bsnm{Raffel}, \binits{C.}},
\bauthor{\bsnm{Shazeer}, \binits{N.}},
\bauthor{\bsnm{Roberts}, \binits{A.}},
\bauthor{\bsnm{Lee}, \binits{K.}},
\bauthor{\bsnm{Narang}, \binits{S.}},
\bauthor{\bsnm{Matena}, \binits{M.}},
\bauthor{\bsnm{Zhou}, \binits{Y.}},
\bauthor{\bsnm{Li}, \binits{W.}},
\bauthor{\bsnm{Liu}, \binits{P.J.}}:
\batitle{Exploring the limits of transfer learning with a unified text-to-text transformer}.
\bjtitle{Journal of Machine Learning Research}
\bvolume{21}(\bissue{140}),
\bfpage{1}--\blpage{67}
(\byear{2020})
\end{barticle}
\endbibitem

\bibitem[\protect\citeauthoryear{Lewis et~al.}{2020}]{lewis_2020_bart}
\begin{bchapter}
\bauthor{\bsnm{Lewis}, \binits{M.}},
\bauthor{\bsnm{Liu}, \binits{Y.}},
\bauthor{\bsnm{Goyal}, \binits{N.}},
\bauthor{\bsnm{Ghazvininejad}, \binits{M.}},
\bauthor{\bsnm{Mohamed}, \binits{A.}},
\bauthor{\bsnm{Levy}, \binits{O.}},
\bauthor{\bsnm{Stoyanov}, \binits{V.}},
\bauthor{\bsnm{Zettlemoyer}, \binits{L.}}:
\bctitle{{BART}: Denoising sequence-to-sequence pre-training for natural language generation, translation, and comprehension}.
In: \beditor{\bsnm{Jurafsky}, \binits{D.}},
\beditor{\bsnm{Chai}, \binits{J.}},
\beditor{\bsnm{Schluter}, \binits{N.}},
\beditor{\bsnm{Tetreault}, \binits{J.}} (eds.)
\bbtitle{Proceedings of the 58th Annual Meeting of the Association for Computational Linguistics},
pp. \bfpage{7871}--\blpage{7880}.
\bpublisher{Association for Computational Linguistics},
\blocation{Online}
(\byear{2020}).
\doiurl{10.18653/v1/2020.acl-main.703} .
\burl{https://aclanthology.org/2020.acl-main.703/}
\end{bchapter}
\endbibitem

\bibitem[\protect\citeauthoryear{Vaswani et~al.}{2017}]{vaswani_2017_transformer}
\begin{bchapter}
\bauthor{\bsnm{Vaswani}, \binits{A.}},
\bauthor{\bsnm{Shazeer}, \binits{N.}},
\bauthor{\bsnm{Parmar}, \binits{N.}},
\bauthor{\bsnm{Uszkoreit}, \binits{J.}},
\bauthor{\bsnm{Jones}, \binits{L.}},
\bauthor{\bsnm{Gomez}, \binits{A.N.}},
\bauthor{\bsnm{Kaiser}, \binits{L.u.}},
\bauthor{\bsnm{Polosukhin}, \binits{I.}}:
\bctitle{Attention is all you need}.
In: \beditor{\bsnm{Guyon}, \binits{I.}},
\beditor{\bsnm{Luxburg}, \binits{U.V.}},
\beditor{\bsnm{Bengio}, \binits{S.}},
\beditor{\bsnm{Wallach}, \binits{H.}},
\beditor{\bsnm{Fergus}, \binits{R.}},
\beditor{\bsnm{Vishwanathan}, \binits{S.}},
\beditor{\bsnm{Garnett}, \binits{R.}} (eds.)
\bbtitle{Advances in Neural Information Processing Systems},
vol. \bseriesno{30}.
\bpublisher{Curran Associates, Inc.}, \blocation{???}
(\byear{2017})
\end{bchapter}
\endbibitem

\bibitem[\protect\citeauthoryear{Rodriguez-Torrealba et~al.}{2022}]{rodriguez_2022_mcq}
\begin{barticle}
\bauthor{\bsnm{Rodriguez-Torrealba}, \binits{R.}},
\bauthor{\bsnm{Garcia-Lopez}, \binits{E.}},
\bauthor{\bsnm{Garcia-Cabot}, \binits{A.}}:
\batitle{End-to-end generation of multiple-choice questions using text-to-text transfer transformer models}.
\bjtitle{Expert Systems with Applications}
\bvolume{208},
\bfpage{118258}
(\byear{2022})
\doiurl{10.1016/j.eswa.2022.118258}
\end{barticle}
\endbibitem

\bibitem[\protect\citeauthoryear{Wang et~al.}{2023}]{wang_2023_dist}
\begin{bchapter}
\bauthor{\bsnm{Wang}, \binits{H.-J.}},
\bauthor{\bsnm{Hsieh}, \binits{K.-Y.}},
\bauthor{\bsnm{Yu}, \binits{H.-C.}},
\bauthor{\bsnm{Tsou}, \binits{J.-C.}},
\bauthor{\bsnm{Shih}, \binits{Y.A.}},
\bauthor{\bsnm{Huang}, \binits{C.-H.}},
\bauthor{\bsnm{Fan}, \binits{Y.-C.}}:
\bctitle{Distractor generation based on {T}ext2{T}ext language models with pseudo {K}ullback-{L}eibler divergence regulation}.
In: \beditor{\bsnm{Rogers}, \binits{A.}},
\beditor{\bsnm{Boyd-Graber}, \binits{J.}},
\beditor{\bsnm{Okazaki}, \binits{N.}} (eds.)
\bbtitle{Findings of the Association for Computational Linguistics: ACL 2023},
pp. \bfpage{12477}--\blpage{12491}.
\bpublisher{Association for Computational Linguistics},
\blocation{Toronto, Canada}
(\byear{2023}).
\doiurl{10.18653/v1/2023.findings-acl.790} .
\burl{https://aclanthology.org/2023.findings-acl.790/}
\end{bchapter}
\endbibitem

\bibitem[\protect\citeauthoryear{Taslimipoor et~al.}{2024}]{taslimipoor_2024_dist}
\begin{bchapter}
\bauthor{\bsnm{Taslimipoor}, \binits{S.}},
\bauthor{\bsnm{Benedetto}, \binits{L.}},
\bauthor{\bsnm{Felice}, \binits{M.}},
\bauthor{\bsnm{Buttery}, \binits{P.}}:
\bctitle{Distractor generation using generative and discriminative capabilities of transformer-based models}.
In: \beditor{\bsnm{Calzolari}, \binits{N.}},
\beditor{\bsnm{Kan}, \binits{M.-Y.}},
\beditor{\bsnm{Hoste}, \binits{V.}},
\beditor{\bsnm{Lenci}, \binits{A.}},
\beditor{\bsnm{Sakti}, \binits{S.}},
\beditor{\bsnm{Xue}, \binits{N.}} (eds.)
\bbtitle{Proceedings of the 2024 Joint International Conference on Computational Linguistics, Language Resources and Evaluation (LREC-COLING 2024)},
pp. \bfpage{5052}--\blpage{5063}.
\bpublisher{ELRA and ICCL},
\blocation{Torino, Italia}
(\byear{2024}).
\burl{https://aclanthology.org/2024.lrec-main.452/}
\end{bchapter}
\endbibitem

\bibitem[\protect\citeauthoryear{Bitew et~al.}{2025}]{bitew_2025_fewshot}
\begin{bchapter}
\bauthor{\bsnm{Bitew}, \binits{S.K.}},
\bauthor{\bsnm{Deleu}, \binits{J.}},
\bauthor{\bsnm{Develder}, \binits{C.}},
\bauthor{\bsnm{Demeester}, \binits{T.}}:
\bctitle{Distractor generation for multiple-choice questions with predictive prompting and large language models}.
In: \beditor{\bsnm{Meo}, \binits{R.}},
\beditor{\bsnm{Silvestri}, \binits{F.}} (eds.)
\bbtitle{Machine Learning and Principles and Practice of Knowledge Discovery in Databases},
pp. \bfpage{48}--\blpage{63}.
\bpublisher{Springer},
\blocation{Cham}
(\byear{2025})
\end{bchapter}
\endbibitem

\bibitem[\protect\citeauthoryear{Lin and Chen}{2024}]{lin_2024_psyco}
\begin{barticle}
\bauthor{\bsnm{Lin}, \binits{Z.}},
\bauthor{\bsnm{Chen}, \binits{H.}}:
\batitle{Investigating the capability of chatgpt for generating multiple-choice reading comprehension items}.
\bjtitle{System}
\bvolume{123},
\bfpage{103344}
(\byear{2024})
\doiurl{10.1016/j.system.2024.103344}
\end{barticle}
\endbibitem

\bibitem[\protect\citeauthoryear{Papineni et~al.}{2002}]{papineni_bleu_2002}
\begin{bchapter}
\bauthor{\bsnm{Papineni}, \binits{K.}},
\bauthor{\bsnm{Roukos}, \binits{S.}},
\bauthor{\bsnm{Ward}, \binits{T.}},
\bauthor{\bsnm{Zhu}, \binits{W.-J.}}:
\bctitle{Bleu: a {Method} for {Automatic} {Evaluation} of {Machine} {Translation}}.
In: \bbtitle{Proceedings of the 40th {Annual} {Meeting} of the {Association} for {Computational} {Linguistics}},
pp. \bfpage{311}--\blpage{318}.
\bpublisher{ACL},
\blocation{Philadelphia, Pennsylvania, USA}
(\byear{2002}).
\doiurl{10.3115/1073083.1073135}
\end{bchapter}
\endbibitem

\bibitem[\protect\citeauthoryear{Lin}{2004}]{lin_rouge_2004}
\begin{bchapter}
\bauthor{\bsnm{Lin}, \binits{C.-Y.}}:
\bctitle{{ROUGE}: {A} {Package} for {Automatic} {Evaluation} of {Summaries}}.
In: \bbtitle{Text {Summarization} {Branches} {Out}},
pp. \bfpage{74}--\blpage{81}.
\bpublisher{ACL},
\blocation{Barcelona, Spain}
(\byear{2004}).
\burl{https://www.aclweb.org/anthology/W04-1013}
\end{bchapter}
\endbibitem

\bibitem[\protect\citeauthoryear{Curto}{2010}]{curto_mthesis_2010}
\begin{botherref}
\oauthor{\bsnm{Curto}, \binits{S.}}:
Automatic generation of multiple-choice tests.
Master's thesis,
Instituto Superior Técnico
(2010).
Dissertation for obtaining the Master Degree in Information Systems and Computer Engineering.
\url{https://fenix.tecnico.ulisboa.pt/departamentos/dei/dissertacao/2353642299631}
\end{botherref}
\endbibitem

\bibitem[\protect\citeauthoryear{dos Santos~Correia et~al.}{2010}]{correia_mcqpt_2010}
\begin{bchapter}
\bauthor{\bsnm{Santos~Correia}, \binits{R.P.}},
\bauthor{\bsnm{Baptista}, \binits{J.}},
\bauthor{\bsnm{Mamede}, \binits{N.}},
\bauthor{\bsnm{Trancoso}, \binits{I.}},
\bauthor{\bsnm{Eskenazi}, \binits{M.}}:
\bctitle{Automatic generation of cloze question distractors}.
In: \bbtitle{Second Language Studies: Acquisition, Learning, Education and Technology (L2WS 2010)},
pp. \bfpage{2}--\blpage{11}
(\byear{2010})
\end{bchapter}
\endbibitem

\bibitem[\protect\citeauthoryear{Leite}{2020}]{leite_2020_msc}
\begin{botherref}
\oauthor{\bsnm{Leite}, \binits{B.}}:
Automatic question generation for the portuguese language.
Master's thesis,
Faculdade de Engenharia da Universidade do Porto
(2020).
Dissertation for obtaining the Integrated Master Degree in Informatics and Computer Engineering.
\url{https://repositorio-aberto.up.pt/handle/10216/128541}
\end{botherref}
\endbibitem

\bibitem[\protect\citeauthoryear{Gon\c{c}alo~Oliveira et~al.}{2023}]{oliveira_mcqpt_2023}
\begin{bchapter}
\bauthor{\bsnm{Gon\c{c}alo~Oliveira}, \binits{H.}},
\bauthor{\bsnm{Caetano}, \binits{I.}},
\bauthor{\bsnm{Matos}, \binits{R.}},
\bauthor{\bsnm{Amaro}, \binits{H.}}:
\bctitle{{Generating and Ranking Distractors for Multiple-Choice Questions in Portuguese}}.
In: \beditor{\bsnm{Sim\~{o}es}, \binits{A.}},
\beditor{\bsnm{Ber\'{o}n}, \binits{M.M.}},
\beditor{\bsnm{Portela}, \binits{F.}} (eds.)
\bbtitle{12th Symposium on Languages, Applications and Technologies (SLATE 2023)}.
\bsertitle{Open Access Series in Informatics (OASIcs)},
vol. \bseriesno{113},
pp. \bfpage{4}--\blpage{149}.
\bpublisher{Schloss Dagstuhl -- Leibniz-Zentrum f{\"u}r Informatik},
\blocation{Dagstuhl, Germany}
(\byear{2023}).
\doiurl{10.4230/OASIcs.SLATE.2023.4} .
\burl{https://drops.dagstuhl.de/entities/document/10.4230/OASIcs.SLATE.2023.4}
\end{bchapter}
\endbibitem

\bibitem[\protect\citeauthoryear{Ghanem et~al.}{2022}]{ghanem_2022_cqg_acl}
\begin{bchapter}
\bauthor{\bsnm{Ghanem}, \binits{B.}},
\bauthor{\bsnm{Lutz~Coleman}, \binits{L.}},
\bauthor{\bsnm{Rivard~Dexter}, \binits{J.}},
\bauthor{\bsnm{Ohe}, \binits{S.}},
\bauthor{\bsnm{Fyshe}, \binits{A.}}:
\bctitle{Question generation for reading comprehension assessment by modeling how and what to ask}.
In: \bbtitle{Findings of the Association for Computational Linguistics: ACL 2022},
pp. \bfpage{2131}--\blpage{2146}.
\bpublisher{Association for Computational Linguistics},
\blocation{Dublin, Ireland}
(\byear{2022}).
\doiurl{10.18653/v1/2022.findings-acl.168} .
\burl{https://aclanthology.org/2022.findings-acl.168}
\end{bchapter}
\endbibitem

\bibitem[\protect\citeauthoryear{Elkins et~al.}{2023}]{elkins_2023_cqg_aied}
\begin{bchapter}
\bauthor{\bsnm{Elkins}, \binits{S.}},
\bauthor{\bsnm{Kochmar}, \binits{E.}},
\bauthor{\bsnm{Serban}, \binits{I.}},
\bauthor{\bsnm{Cheung}, \binits{J.C.K.}}:
\bctitle{How useful are educational questions generated by large language models?}
In: \beditor{\bsnm{Wang}, \binits{N.}},
\beditor{\bsnm{Rebolledo-Mendez}, \binits{G.}},
\beditor{\bsnm{Dimitrova}, \binits{V.}},
\beditor{\bsnm{Matsuda}, \binits{N.}},
\beditor{\bsnm{Santos}, \binits{O.C.}} (eds.)
\bbtitle{Artificial Intelligence in Education. Posters and Late Breaking Results, Workshops and Tutorials, Industry and Innovation Tracks, Practitioners, Doctoral Consortium and Blue Sky},
pp. \bfpage{536}--\blpage{542}.
\bpublisher{Springer},
\blocation{Cham}
(\byear{2023})
\end{bchapter}
\endbibitem

\bibitem[\protect\citeauthoryear{Krathwohl}{2002}]{bloom_2002_revised}
\begin{barticle}
\bauthor{\bsnm{Krathwohl}, \binits{D.R.}}:
\batitle{A revision of bloom's taxonomy: An overview}.
\bjtitle{Theory Into Practice}
\bvolume{41}(\bissue{4}),
\bfpage{212}--\blpage{218}
(\byear{2002})
\doiurl{10.1207/s15430421tip4104\_2}
\end{barticle}
\endbibitem

\bibitem[\protect\citeauthoryear{Zhao et~al.}{2022}]{zhao_2022_cqg_acl}
\begin{bchapter}
\bauthor{\bsnm{Zhao}, \binits{Z.}},
\bauthor{\bsnm{Hou}, \binits{Y.}},
\bauthor{\bsnm{Wang}, \binits{D.}},
\bauthor{\bsnm{Yu}, \binits{M.}},
\bauthor{\bsnm{Liu}, \binits{C.}},
\bauthor{\bsnm{Ma}, \binits{X.}}:
\bctitle{Educational question generation of children storybooks via question type distribution learning and event-centric summarization}.
In: \bbtitle{Proceedings of the 60th Annual Meeting of the Association for Computational Linguistics (Volume 1: Long Papers)},
pp. \bfpage{5073}--\blpage{5085}.
\bpublisher{Association for Computational Linguistics},
\blocation{Dublin, Ireland}
(\byear{2022}).
\doiurl{10.18653/v1/2022.acl-long.348} .
\burl{https://aclanthology.org/2022.acl-long.348}
\end{bchapter}
\endbibitem

\bibitem[\protect\citeauthoryear{Leite and Lopes~Cardoso}{2023}]{leite_2023_cqg_aied}
\begin{bchapter}
\bauthor{\bsnm{Leite}, \binits{B.}},
\bauthor{\bsnm{Lopes~Cardoso}, \binits{H.}}:
\bctitle{Towards enriched controllability for educational question generation}.
In: \beditor{\bsnm{Wang}, \binits{N.}},
\beditor{\bsnm{Rebolledo-Mendez}, \binits{G.}},
\beditor{\bsnm{Matsuda}, \binits{N.}},
\beditor{\bsnm{Santos}, \binits{O.C.}},
\beditor{\bsnm{Dimitrova}, \binits{V.}} (eds.)
\bbtitle{Artificial Intelligence in Education},
pp. \bfpage{786}--\blpage{791}.
\bpublisher{Springer},
\blocation{Cham}
(\byear{2023})
\end{bchapter}
\endbibitem

\bibitem[\protect\citeauthoryear{Leite and Lopes~Cardoso}{2024}]{leite_2024_fcqg_csedu}
\begin{bchapter}
\bauthor{\bsnm{Leite}, \binits{B.}},
\bauthor{\bsnm{Lopes~Cardoso}, \binits{H.}}:
\bctitle{On Few-Shot Prompting for Controllable Question-Answer Generation in Narrative Comprehension}.
In: \bbtitle{Proceedings of the 16th International Conference on Computer Supported Education - Volume 2: CSEDU},
pp. \bfpage{63}--\blpage{74}.
\bpublisher{SciTePress}, \blocation{???}
(\byear{2024}).
\doiurl{10.5220/0012623800003693} .
\bcomment{INSTICC}
\end{bchapter}
\endbibitem

\bibitem[\protect\citeauthoryear{Li and Zhang}{2024}]{zhang_2024_plans}
\begin{bchapter}
\bauthor{\bsnm{Li}, \binits{K.}},
\bauthor{\bsnm{Zhang}, \binits{Y.}}:
\bctitle{Planning first, question second: An {LLM}-guided method for controllable question generation}.
In: \beditor{\bsnm{Ku}, \binits{L.-W.}},
\beditor{\bsnm{Martins}, \binits{A.}},
\beditor{\bsnm{Srikumar}, \binits{V.}} (eds.)
\bbtitle{Findings of the Association for Computational Linguistics: ACL 2024},
pp. \bfpage{4715}--\blpage{4729}.
\bpublisher{Association for Computational Linguistics},
\blocation{Bangkok, Thailand}
(\byear{2024}).
\doiurl{10.18653/v1/2024.findings-acl.280} .
\burl{https://aclanthology.org/2024.findings-acl.280}
\end{bchapter}
\endbibitem

\bibitem[\protect\citeauthoryear{Gao et~al.}{2019}]{yifangao_2019_dqg}
\begin{bchapter}
\bauthor{\bsnm{Gao}, \binits{Y.}},
\bauthor{\bsnm{Bing}, \binits{L.}},
\bauthor{\bsnm{Chen}, \binits{W.}},
\bauthor{\bsnm{Lyu}, \binits{M.}},
\bauthor{\bsnm{King}, \binits{I.}}:
\bctitle{Difficulty controllable generation of reading comprehension questions}.
In: \bbtitle{Proceedings of the Twenty-Eighth International Joint Conference on Artificial Intelligence, {IJCAI-19}},
pp. \bfpage{4968}--\blpage{4974}.
\bpublisher{International Joint Conferences on Artificial Intelligence Organization}, \blocation{???}
(\byear{2019}).
\doiurl{10.24963/ijcai.2019/690} .
\burl{https://doi.org/10.24963/ijcai.2019/690}
\end{bchapter}
\endbibitem

\bibitem[\protect\citeauthoryear{Kumar et~al.}{2019}]{kumar_2019_dcqg}
\begin{bchapter}
\bauthor{\bsnm{Kumar}, \binits{V.}},
\bauthor{\bsnm{Hua}, \binits{Y.}},
\bauthor{\bsnm{Ramakrishnan}, \binits{G.}},
\bauthor{\bsnm{Qi}, \binits{G.}},
\bauthor{\bsnm{Gao}, \binits{L.}},
\bauthor{\bsnm{Li}, \binits{Y.-F.}}:
\bctitle{Difficulty-controllable multi-hop question generation from knowledge graphs}.
In: \bbtitle{The Semantic Web – ISWC 2019: 18th International Semantic Web Conference, Auckland, New Zealand, October 26–30, 2019, Proceedings, Part I},
pp. \bfpage{382}--\blpage{398}.
\bpublisher{Springer},
\blocation{Berlin, Heidelberg}
(\byear{2019}).
\doiurl{10.1007/978-3-030-30793-6_22}
\end{bchapter}
\endbibitem

\bibitem[\protect\citeauthoryear{Bi et~al.}{2021}]{bi_2021_dcqg}
\begin{botherref}
\oauthor{\bsnm{Bi}, \binits{S.}},
\oauthor{\bsnm{Cheng}, \binits{X.}},
\oauthor{\bsnm{Li}, \binits{Y.-F.}},
\oauthor{\bsnm{Qu}, \binits{L.}},
\oauthor{\bsnm{Shen}, \binits{S.}},
\oauthor{\bsnm{Qi}, \binits{G.}},
\oauthor{\bsnm{Pan}, \binits{L.}},
\oauthor{\bsnm{Jiang}, \binits{Y.}}:
Simple or complex? complexity-controllable question generation with soft templates and deep mixture of experts model.
arXiv preprint arXiv:2110.06560
(2021)
\end{botherref}
\endbibitem

\bibitem[\protect\citeauthoryear{Cheng et~al.}{2021}]{cheng_2021_dcqg}
\begin{bchapter}
\bauthor{\bsnm{Cheng}, \binits{Y.}},
\bauthor{\bsnm{Li}, \binits{S.}},
\bauthor{\bsnm{Liu}, \binits{B.}},
\bauthor{\bsnm{Zhao}, \binits{R.}},
\bauthor{\bsnm{Li}, \binits{S.}},
\bauthor{\bsnm{Lin}, \binits{C.}},
\bauthor{\bsnm{Zheng}, \binits{Y.}}:
\bctitle{Guiding the growth: Difficulty-controllable question generation through step-by-step rewriting}.
In: \bbtitle{Proceedings of the 59th Annual Meeting of the Association for Computational Linguistics and the 11th International Joint Conference on Natural Language Processing (Volume 1: Long Papers)},
pp. \bfpage{5968}--\blpage{5978}.
\bpublisher{Association for Computational Linguistics},
\blocation{Online}
(\byear{2021}).
\doiurl{10.18653/v1/2021.acl-long.465} .
\burl{https://aclanthology.org/2021.acl-long.465}
\end{bchapter}
\endbibitem

\bibitem[\protect\citeauthoryear{Uto et~al.}{2023}]{uto_2023_dif}
\begin{bchapter}
\bauthor{\bsnm{Uto}, \binits{M.}},
\bauthor{\bsnm{Tomikawa}, \binits{Y.}},
\bauthor{\bsnm{Suzuki}, \binits{A.}}:
\bctitle{Difficulty-controllable neural question generation for reading comprehension using item response theory}.
In: \beditor{\bsnm{Kochmar}, \binits{E.}},
\beditor{\bsnm{Burstein}, \binits{J.}},
\beditor{\bsnm{Horbach}, \binits{A.}},
\beditor{\bsnm{Laarmann-Quante}, \binits{R.}},
\beditor{\bsnm{Madnani}, \binits{N.}},
\beditor{\bsnm{Tack}, \binits{A.}},
\beditor{\bsnm{Yaneva}, \binits{V.}},
\beditor{\bsnm{Yuan}, \binits{Z.}},
\beditor{\bsnm{Zesch}, \binits{T.}} (eds.)
\bbtitle{Proceedings of the 18th Workshop on Innovative Use of NLP for Building Educational Applications (BEA 2023)},
pp. \bfpage{119}--\blpage{129}.
\bpublisher{Association for Computational Linguistics},
\blocation{Toronto, Canada}
(\byear{2023}).
\doiurl{10.18653/v1/2023.bea-1.10} .
\burl{https://aclanthology.org/2023.bea-1.10}
\end{bchapter}
\endbibitem

\bibitem[\protect\citeauthoryear{Tomikawa et~al.}{2024}]{tomikawa_2024_adpative}
\begin{barticle}
\bauthor{\bsnm{Tomikawa}, \binits{Y.}},
\bauthor{\bsnm{Suzuki}, \binits{A.}},
\bauthor{\bsnm{Uto}, \binits{M.}}:
\batitle{Adaptive question–answer generation with difficulty control using item response theory and pretrained transformer models}.
\bjtitle{IEEE Transactions on Learning Technologies}
\bvolume{17},
\bfpage{2240}--\blpage{2252}
(\byear{2024})
\doiurl{10.1109/TLT.2024.3491801}
\end{barticle}
\endbibitem

\bibitem[\protect\citeauthoryear{Lin et~al.}{2015}]{lin_2015_dif}
\begin{bchapter}
\bauthor{\bsnm{Lin}, \binits{C.}},
\bauthor{\bsnm{Liu}, \binits{D.}},
\bauthor{\bsnm{Pang}, \binits{W.}},
\bauthor{\bsnm{Apeh}, \binits{E.}}:
\bctitle{Automatically predicting quiz difficulty level using similarity measures}.
In: \bbtitle{Proceedings of the 8th International Conference on Knowledge Capture}.
\bsertitle{K-CAP 2015}.
\bpublisher{Association for Computing Machinery},
\blocation{New York, NY, USA}
(\byear{2015}).
\doiurl{10.1145/2815833.2815842} .
\burl{https://doi.org/10.1145/2815833.2815842}
\end{bchapter}
\endbibitem

\bibitem[\protect\citeauthoryear{Kurdi et~al.}{2016}]{kurdi_2016_dif}
\begin{bchapter}
\bauthor{\bsnm{Kurdi}, \binits{G.}},
\bauthor{\bsnm{Parsia}, \binits{B.}},
\bauthor{\bsnm{Sattler}, \binits{U.}}:
\bctitle{An experimental evaluation of automatically generated multiple choice questions from ontologies}.
In: \bbtitle{OWL: Experiences And Directions--reasoner Evaluation},
pp. \bfpage{24}--\blpage{39}.
\bpublisher{Springer}, \blocation{???}
(\byear{2016})
\end{bchapter}
\endbibitem

\bibitem[\protect\citeauthoryear{Susanti et~al.}{2017}]{susanti_2017_cid}
\begin{barticle}
\bauthor{\bsnm{Susanti}, \binits{Y.}},
\bauthor{\bsnm{Tokunaga}, \binits{T.}},
\bauthor{\bsnm{Nishikawa}, \binits{H.}},
\bauthor{\bsnm{Obari}, \binits{H.}}:
\batitle{Controlling item difficulty for automatic vocabulary question generation}.
\bjtitle{Research and practice in technology enhanced learning}
\bvolume{12}(\bissue{1}),
\bfpage{1}--\blpage{16}
(\byear{2017})
\end{barticle}
\endbibitem

\bibitem[\protect\citeauthoryear{Tomikawa and Uto}{2024}]{tomikawa_2024_difmcqg}
\begin{bchapter}
\bauthor{\bsnm{Tomikawa}, \binits{Y.}},
\bauthor{\bsnm{Uto}, \binits{M.}}:
\bctitle{Difficulty-controllable multiple-choice question generation for reading comprehension using item response theory}.
In: \beditor{\bsnm{Olney}, \binits{A.M.}},
\beditor{\bsnm{Chounta}, \binits{I.-A.}},
\beditor{\bsnm{Liu}, \binits{Z.}},
\beditor{\bsnm{Santos}, \binits{O.C.}},
\beditor{\bsnm{Bittencourt}, \binits{I.I.}} (eds.)
\bbtitle{Artificial Intelligence in Education. Posters and Late Breaking Results, Workshops and Tutorials, Industry and Innovation Tracks, Practitioners, Doctoral Consortium and Blue Sky},
pp. \bfpage{312}--\blpage{320}.
\bpublisher{Springer},
\blocation{Cham}
(\byear{2024})
\end{bchapter}
\endbibitem

\bibitem[\protect\citeauthoryear{Ray}{2023}]{ray_principles_2023}
\begin{barticle}
\bauthor{\bsnm{Ray}, \binits{P.P.}}:
\batitle{Chatgpt: A comprehensive review on background, applications, key challenges, bias, ethics, limitations and future scope}.
\bjtitle{Internet of Things and Cyber-Physical Systems}
\bvolume{3},
\bfpage{121}--\blpage{154}
(\byear{2023})
\doiurl{10.1016/j.iotcps.2023.04.003}
\end{barticle}
\endbibitem

\bibitem[\protect\citeauthoryear{Heston and Khun}{2023}]{heston_interactiveprompt_2023}
\begin{barticle}
\bauthor{\bsnm{Heston}, \binits{T.F.}},
\bauthor{\bsnm{Khun}, \binits{C.}}:
\batitle{Prompt engineering in medical education}.
\bjtitle{International Medical Education}
\bvolume{2}(\bissue{3}),
\bfpage{198}--\blpage{205}
(\byear{2023})
\doiurl{10.3390/ime2030019}
\end{barticle}
\endbibitem

\bibitem[\protect\citeauthoryear{Wu et~al.}{2022}]{wu_2022_divide_nlp_tasks}
\begin{bchapter}
\bauthor{\bsnm{Wu}, \binits{T.}},
\bauthor{\bsnm{Terry}, \binits{M.}},
\bauthor{\bsnm{Cai}, \binits{C.J.}}:
\bctitle{Ai chains: Transparent and controllable human-ai interaction by chaining large language model prompts}.
In: \bbtitle{Proceedings of the 2022 CHI Conference on Human Factors in Computing Systems}.
\bsertitle{CHI '22}.
\bpublisher{Association for Computing Machinery},
\blocation{New York, NY, USA}
(\byear{2022}).
\doiurl{10.1145/3491102.3517582} .
\burl{https://doi.org/10.1145/3491102.3517582}
\end{bchapter}
\endbibitem

\bibitem[\protect\citeauthoryear{Piau et~al.}{2024}]{piau_2024_ptt5v2}
\begin{bchapter}
\bauthor{\bsnm{Piau}, \binits{M.}},
\bauthor{\bsnm{Lotufo}, \binits{R.}},
\bauthor{\bsnm{Nogueira}, \binits{R.}}:
\bctitle{ptt5-v2: A closer look at continued pretraining of t5 models for the portuguese language}.
In: \bbtitle{Brazilian Conference on Intelligent Systems},
pp. \bfpage{324}--\blpage{338}
(\byear{2024}).
\bcomment{Springer}
\end{bchapter}
\endbibitem

\bibitem[\protect\citeauthoryear{Leite et~al.}{2024}]{leite_2024_fairy_pt}
\begin{bchapter}
\bauthor{\bsnm{Leite}, \binits{B.}},
\bauthor{\bsnm{Os{\'o}rio}, \binits{T.F.}},
\bauthor{\bsnm{Cardoso}, \binits{H.L.}}:
\bctitle{Fairytaleqa translated: Enabling educational question and answer generation in less-resourced languages}.
In: \beditor{\bsnm{Ferreira~Mello}, \binits{R.}},
\beditor{\bsnm{Rummel}, \binits{N.}},
\beditor{\bsnm{Jivet}, \binits{I.}},
\beditor{\bsnm{Pishtari}, \binits{G.}},
\beditor{\bsnm{Ruip{\'e}rez~Valiente}, \binits{J.A.}} (eds.)
\bbtitle{Technology Enhanced Learning for Inclusive and Equitable Quality Education},
pp. \bfpage{222}--\blpage{236}.
\bpublisher{Springer},
\blocation{Cham}
(\byear{2024})
\end{bchapter}
\endbibitem

\bibitem[\protect\citeauthoryear{Haegeman}{2007}]{haegeman_2007_subject}
\begin{botherref}
\oauthor{\bsnm{Haegeman}, \binits{L.}}:
Subject omission in present-day written english.
Rivista di Grammatica Generativa, vol. 32 (2007), p. 91-124
(2007)
\end{botherref}
\endbibitem

\bibitem[\protect\citeauthoryear{Haladyna}{2004}]{haladyna_2004_mcq_validation}
\begin{bbook}
\bauthor{\bsnm{Haladyna}, \binits{T.M.}}:
\bbtitle{Developing and Validating Multiple-choice Test Items}.
\bpublisher{Routledge}, \blocation{???}
(\byear{2004})
\end{bbook}
\endbibitem

\bibitem[\protect\citeauthoryear{Crocker and Algina}{1986}]{crocker_86_ctt}
\begin{bbook}
\bauthor{\bsnm{Crocker}, \binits{L.}},
\bauthor{\bsnm{Algina}, \binits{J.}}:
\bbtitle{Introduction to Classical and Modern Test Theory.}
\bpublisher{ERIC}, \blocation{???}
(\byear{1986})
\end{bbook}
\endbibitem

\bibitem[\protect\citeauthoryear{Martinkov{\'a} and Drabinov{\'a}}{2018}]{martinkova_2018_shinyitemanalysis}
\begin{botherref}
\oauthor{\bsnm{Martinkov{\'a}}, \binits{P.}},
\oauthor{\bsnm{Drabinov{\'a}}, \binits{A.}}:
Shinyitemanalysis for teaching psychometrics and to enforce routine analysis of educational tests.
R Journal
\textbf{10}(2)
(2018)
\end{botherref}
\endbibitem

\bibitem[\protect\citeauthoryear{Rodrigues et~al.}{2023}]{rodrigues_2023_albertina}
\begin{bchapter}
\bauthor{\bsnm{Rodrigues}, \binits{J.}},
\bauthor{\bsnm{Gomes}, \binits{L.}},
\bauthor{\bsnm{Silva}, \binits{J.}},
\bauthor{\bsnm{Branco}, \binits{A.}},
\bauthor{\bsnm{Santos}, \binits{R.}},
\bauthor{\bsnm{Cardoso}, \binits{H.L.}},
\bauthor{\bsnm{Os{\'o}rio}, \binits{T.}}:
\bctitle{Advancing neural encoding of portuguese with transformer albertina pt}.
In: \bbtitle{EPIA Conference on Artificial Intelligence},
pp. \bfpage{441}--\blpage{453}
(\byear{2023}).
\bcomment{Springer}
\end{bchapter}
\endbibitem

\bibitem[\protect\citeauthoryear{Scaria et~al.}{2024}]{scaria_2024_aied}
\begin{bchapter}
\bauthor{\bsnm{Scaria}, \binits{N.}},
\bauthor{\bsnm{Dharani~Chenna}, \binits{S.}},
\bauthor{\bsnm{Subramani}, \binits{D.}}:
\bctitle{Automated educational question generation at different bloom's skill levels using large language models: Strategies and evaluation}.
In: \beditor{\bsnm{Olney}, \binits{A.M.}},
\beditor{\bsnm{Chounta}, \binits{I.-A.}},
\beditor{\bsnm{Liu}, \binits{Z.}},
\beditor{\bsnm{Santos}, \binits{O.C.}},
\beditor{\bsnm{Bittencourt}, \binits{I.I.}} (eds.)
\bbtitle{Artificial Intelligence in Education},
pp. \bfpage{165}--\blpage{179}.
\bpublisher{Springer},
\blocation{Cham}
(\byear{2024})
\end{bchapter}
\endbibitem

\bibitem[\protect\citeauthoryear{Paris and Paris}{2003}]{paris_2003_narrative}
\begin{barticle}
\bauthor{\bsnm{Paris}, \binits{A.H.}},
\bauthor{\bsnm{Paris}, \binits{S.G.}}:
\batitle{Assessing narrative comprehension in young children}.
\bjtitle{Reading Research Quarterly}
\bvolume{38}(\bissue{1}),
\bfpage{36}--\blpage{76}
(\byear{2003})
\end{barticle}
\endbibitem

\bibitem[\protect\citeauthoryear{Alonzo et~al.}{2009}]{alonzo_2009_narrative}
\begin{barticle}
\bauthor{\bsnm{Alonzo}, \binits{J.}},
\bauthor{\bsnm{Basaraba}, \binits{D.}},
\bauthor{\bsnm{Tindal}, \binits{G.}},
\bauthor{\bsnm{Carriveau}, \binits{R.S.}}:
\batitle{They read, but how well do they understand? an empirical look at the nuances of measuring reading comprehension}.
\bjtitle{Assessment for Effective Intervention}
\bvolume{35}(\bissue{1}),
\bfpage{34}--\blpage{44}
(\byear{2009})
\end{barticle}
\endbibitem

\end{thebibliography}

\begin{appendices}

\section{Narrative Elements}\label{sec:appendix_narratives}

The narrative elements targeted for control are:
\begin{itemize}
    \item \textbf{Character}: Identifying or describing characters' identities or traits (e.g., ``Who...?'' );
    \item \textbf{Setting}: Focusing on the time and place of events, often starting with ``Where...?'' or ``When...'';
    \item \textbf{Action}: Related to the actions of characters;
    \item \textbf{Feeling}: Exploring emotional states or reactions (e.g., ``How did/does X feel?'');
    \item \textbf{Causal relationship}: Addressing cause-and-effect (e.g., ``Why...?'' or ``What caused/made X?'');
\end{itemize}

These narrative elements are based on the FairytaleQA~\citep{xu_2022_fairytaleqa} dataset annotations for each question. The annotations are grounded in evidence-based frameworks~\citep{paris_2003_narrative,alonzo_2009_narrative}

\section{Expert Evaluation Form}\label{sec:expert_eval_form}

Figure~\ref{fig:inquiry_form} shows the inquiry template for expert evaluation.

\begin{figure}[htbp] 
    \centering
    \fbox{
        \begin{minipage}{0.9\textwidth}
            {
            \textbf{Consider the following text:}
            
            \textit{Once upon a time...}  

            \vspace{0.4cm}
            \textbf{Consider the following \textit{wh}-question about the text:}
            
            \textit{What is \ldots ?}  

            \vspace{0.4cm}
            \textbf{Evaluation Metrics}  

            \begin{itemize}
                \item \textbf{Well-formedness:} Is the question well-formed? (Select one)
                \begin{itemize}
                    \item[$\square$] The question is well formulated and has no errors.
                    \item[$\square$] The question is well formulated and has no errors, but is written in the Brazilian Portuguese variant.
                    \item[$\square$] The question is not well formulated as it contains orthographic or punctuation errors.
                    \item[$\square$] The question is not well formulated as it contains grammatical errors.
                    \item[$\square$] The question is not well formulated because it contains semantic errors, for example ambiguity, lack of clarity, or inappropriate terms.
                    \item[$\square$] The question is poorly formulated as it contains several of the errors listed.
                \end{itemize}
            \end{itemize}

            \begin{itemize}
                \item \textbf{Narrative Alignment:} What predominant narrative aspect does the question address? (Select one)
                \begin{itemize}
                    \item[$\square$] Characters. Example: ``Who...?''
                    \item[$\square$] Feelings. Example: ``What was the feeling...?''
                    \item[$\square$] Setting. Example: ``Where...?'', or ''When...?''
                    \item[$\square$] Action. Example: ``What...?'', or ``How...?''
                    \item[$\square$] Causal relationship: Example: ``Why...?''
                \end{itemize}
            \end{itemize}

            \begin{itemize}
                \item \textbf{Answerability 1:} In the text you have read, do you think there is an answer to the question? (Select one)
                \begin{itemize}
                    \item[$\square$] Yes, the answer is in the text (either explicitly or implicitly).
                    \item[$\square$] No, the answer is not in the text.
                \end{itemize}
            \end{itemize}

            \vspace{0.4cm}
            Now, consider the same question again, but this time presented with multiple-choice options:
            \begin{center}
                (A) Option... \quad (B) Option... \quad (C) Option... \quad (D) Option...
            \end{center}

            \vspace{0.4cm}

            \begin{itemize}
                \item \textbf{Answerability 2:} Of the options above, do any of them correspond to the correct answer? (Select one or more options.)
                \begin{center}
                    (A) Option... \quad (B) Option... \quad (C) Option... \quad (D) Option...
                \end{center}
            \end{itemize}

            \begin{itemize}
                \item \textbf{Distractor Plausability:} In the previous question, after selecting the correct answer(s), you should have detected incorrect options (distractors). How do you rate the plausibility of these incorrect options?
                \begin{center}
                (5 Likert-Scale)
                \end{center}
            \end{itemize}
            
            \begin{itemize}
                \item \textbf{Difficulty:} How would you rate the difficulty of the multiple-choice question for a child aged around 8?
                \begin{center}
                (5 Likert-Scale)
                \end{center}
            \end{itemize}

            \begin{itemize}
                \item \textbf{Observations:} If you have any additional comments on the question, please write them below. For example, if you would like to justify one (or some) of your answers. (Free text)
            \end{itemize}
            
            } 
        \end{minipage}
    }
    \caption{Template example of an inquiry form used for expert evaluation.}
    \label{fig:inquiry_form}
\end{figure}

\section{Full Prompt}\label{sec:full_prompt_image}

\begin{figure}[!ht]
    \centering
    \includegraphics[width=\textwidth]{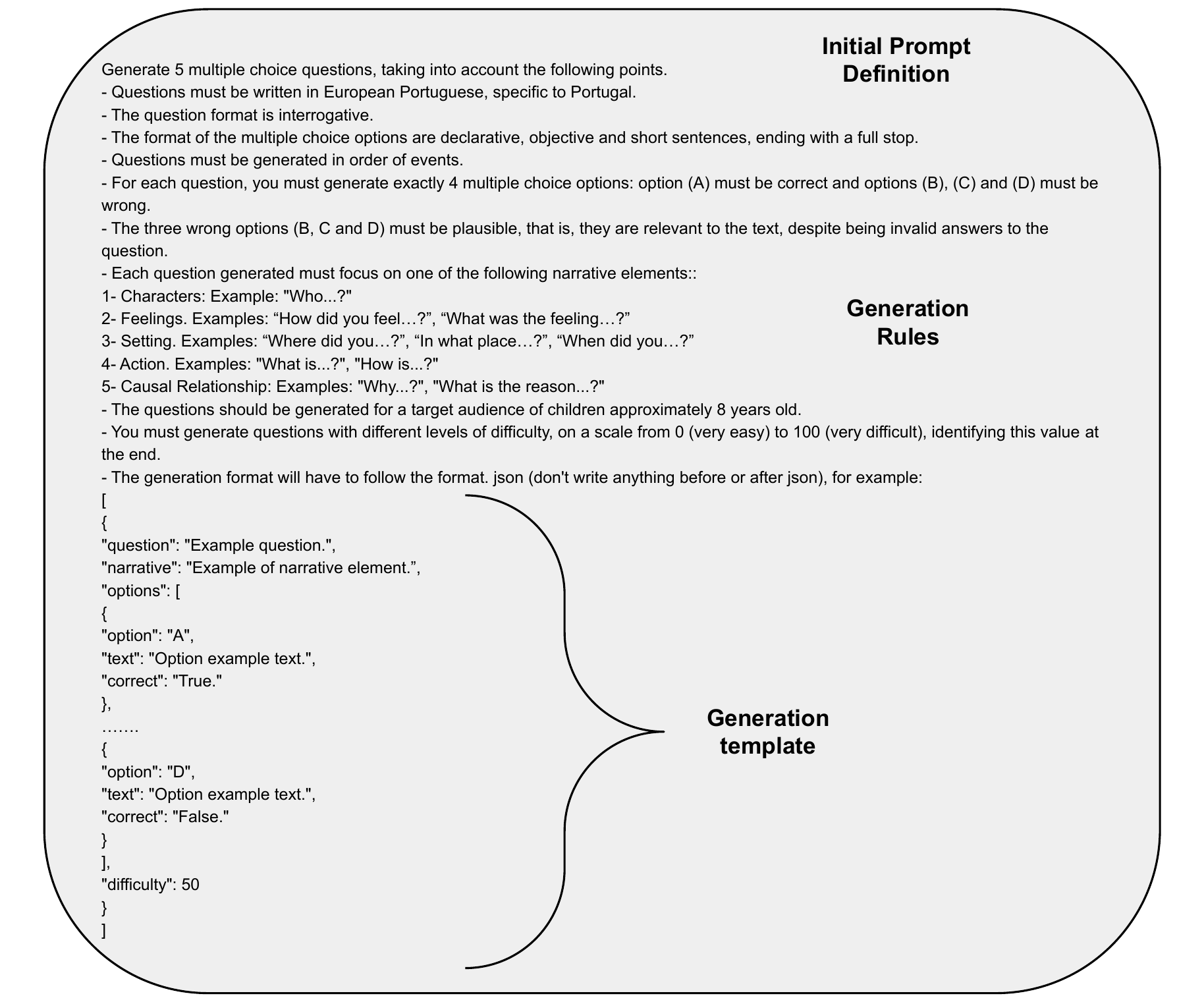}
    \caption{Prompt for one-step MCQ generation.}
    \label{fig:prompt_one_step}
\end{figure}

Figure~\ref{fig:prompt_one_step} shows the final prompt for one-step MCQ generation.


\end{appendices}



\end{document}